\documentclass[runningheads]{llncs}

\usepackage{eccv}

\usepackage{eccvabbrv}

\usepackage{graphicx}
\usepackage{booktabs}
\usepackage{indentfirst} 
\usepackage{algorithm}
\usepackage{algpseudocode}
\usepackage{enumitem}
\usepackage{marvosym}
\usepackage[accsupp]{axessibility}  

\usepackage{hyperref}

\usepackage{orcidlink}

\begin{document}

\title{Omni6D: Large-Vocabulary 3D Object Dataset for Category-Level 6D Object Pose Estimation} 

\titlerunning{Omni6D}

\author{Mengchen Zhang\inst{1,2} \and
Tong Wu\inst{3}\and
Tai Wang\inst{2}\and
Tengfei Wang\inst{2}\and
Ziwei Liu\inst{4}\and
Dahua Lin\inst{2,3}}
\authorrunning{M.~Zhang et al.}

\institute{Zhejiang University, Zhejiang, China\\ \and
Shanghai Artificial Intelligence Laboratory, Shanghai, China\\ \and
The Chinese University of Hong Kong, Hong Kong SAR\\ \and
Nanyang Technological University, Singapore\\
\email{\{zhangmengchen,wangtai,wangtengfei\}@pjlab.org.cn, 
 \{wt020,dhlin\}@ie.cuhk.edu.hk, ziwei.liu@ntu.edu.sg}
}
\maketitle

\setcounter{footnote}{0}

\begin{center}
    \captionsetup{type=figure}
    \includegraphics[width=1\textwidth]{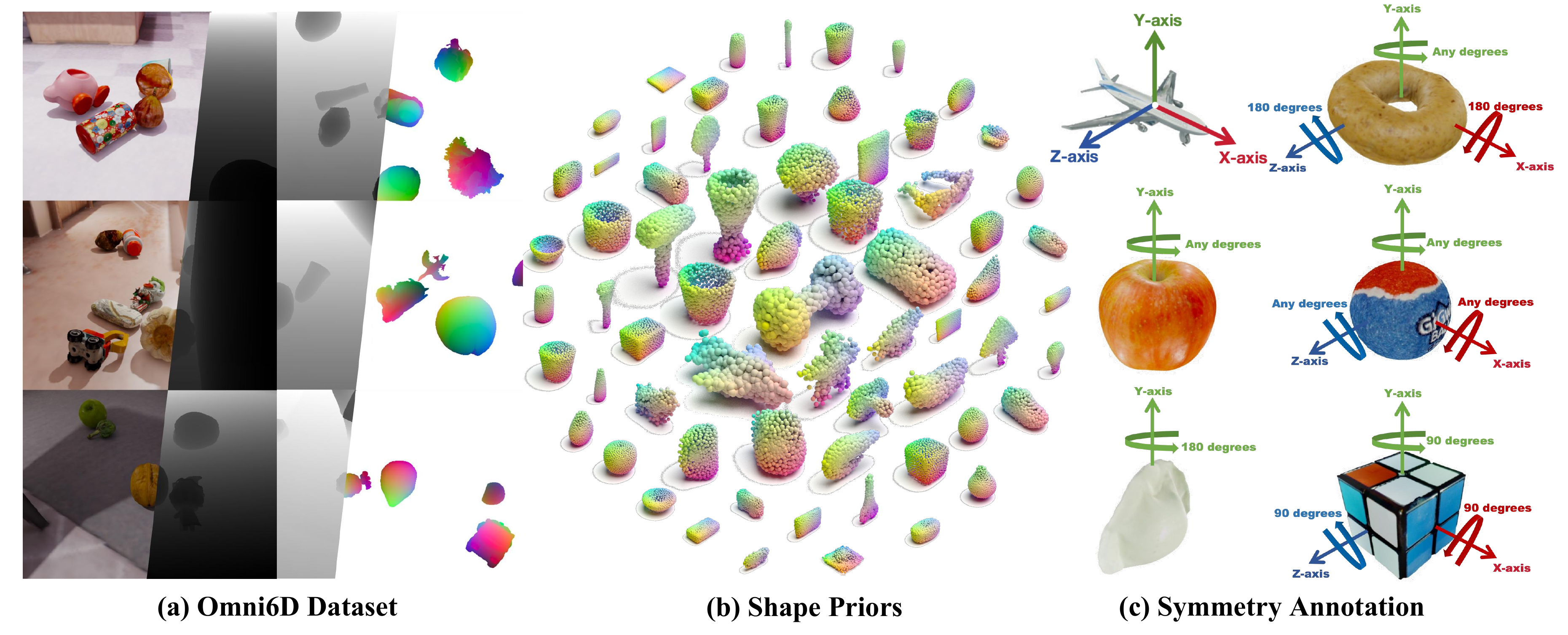}
    \caption{\textbf{Omni6D is a dataset for 6D object pose and size estimation with large vocabulary categories and rich annotations.} \textbf{(a)} showcases ground truth of RGB image, depth map and NOCS map. \textbf{(b)} presents shape priors derived from a variational autoencoder~\cite{Shape_prior} with adjusted canonical poses. \textbf{(c)} provides examples of the rotational symmetry of objects we have annotated, indicating the multiples of angles by which the shape remains unchanged when rotated around the xyz axes.}
    \label{fig:teaser}
\end{center}

\begin{abstract}
6D object pose estimation aims at determining an object's translation, rotation, and scale, typically from a single RGBD image. Recent advancements have expanded this estimation from instance-level to category-level, allowing models to generalize across unseen instances within the same category.
However, this generalization is limited by the narrow range of categories covered by existing datasets, such as NOCS, which also tend to overlook common real-world challenges like occlusion.
To tackle these challenges, we introduce \textbf{Omni6D}, a comprehensive RGBD dataset featuring a wide range of categories and varied backgrounds, elevating the task to a more realistic context. 
\textbf{1)} The dataset comprises an extensive spectrum of \textbf{166} categories, \textbf{4688} instances adjusted to the canonical pose, and over \textbf{0.8 million} captures, significantly broadening the scope for evaluation.
\textbf{2)} We introduce a symmetry-aware metric and conduct systematic benchmarks of existing algorithms on Omni6D, offering a thorough exploration of new challenges and insights. 
\textbf{3)} Additionally, we propose an effective fine-tuning approach that adapts models from previous datasets to our extensive vocabulary setting. 
We believe this initiative will pave the way for new insights and substantial progress in both the industrial and academic fields, pushing forward the boundaries of general 6D pose estimation. 
\end{abstract}
\section{Introduction}
\label{sec:intro}
6D pose estimation aims at predicting the position, orientation, and size of objects in a 3D space using RGB~(D) images, enabling various applications such as augmented/virtual reality~\cite{VR1,VR2}, robot manipulation~\cite{robot1,robot2}, and scene understanding~\cite{understand1,understand2, Voxurf}.

Early instance-level pose estimation approaches~\cite{DPOD,HybridPose,PoseCNN,Occlusion,GDR-Net} typically involve providing instance CAD models and predicting poses of instances that were seen during training, restricting the generalization to unseen objects.
In contrast, recent research has shifted towards category-level 6D object pose estimation~\cite{NOCS,6-PACK,SPD, DualPoseNet,SGPA,FS-Net,reponet,CATRE,nas,GPV, RBP, HS, CenterSnap, shapo, FSD, ss1, ss2}, which learns category prior from a large number of instances within a category, allowing for pose estimation of new instances within the samze category without the need for CAD models. 
By learning on a diverse range of categories, category-level approaches could be a more versatile solution for 6D pose estimation in real-world scenarios. 

However, most existing datasets~\cite{inerf,reponet,NOCS} are limited to a small number of object categories, typically less than 10, as shown in Tab.~\ref{tab:dataset}, hindering their practical applicability to complex scenes. 

To overcome the limitations in previous category-level 6D pose estimation datasets, such as limited category numbers, lack of instance diversity within categories, and overly simplistic scenes, this paper presents a novel category-level dataset dubbed \textbf{\textit{Omni6D}} for 6D pose estimation. \textbf{\textit{Omni6D}} significantly extends the number of object categories to \textbf{166}, and includes \textbf{4,688} real-scanned and well-annotated instance objects with a diverse range of shapes, sizes, and textures. The constructed benchmark includes \textbf{0.8M} images featuring complex scenes with various occlusions, changing lighting conditions, complex backgrounds, and varying viewpoints. For each scene, we provide the rendered image, depth map, NOCS map, and instance mask. Also, considering the widespread rotational symmetry in objects, we examine three types of rotational invariance where an object maintains its original shape under following rotations: any degrees (Sym-1), multiples of 90 degrees (Sym-2) and 180 degrees (Sym-3).  Additionally, we introduce a symmetry-aware metric to specifically address rotational invariance. 
Every object in Omni6D is adjusted to the canonical pose and annotated with rotational symmetry around three axes.

Including a broader range of categories, our dataset offers a more comprehensive and challenging evaluation benchmark for category-level 6D object pose estimation. Utilizing Omni6D, we train and analyze existing algorithms, initiating a profound exploration of the challenges and vital elements involved in category-level estimation within large-vocabulary categories. Additionally, we assess these algorithms' capability to generalize across categories, and carry out a category-wise analysis. Experiments show that our dataset presents a more challenging benchmark for 6D pose estimation, highlighting the need for more robust and generalized pose estimation approaches. As an initial attempt, we present a finetuning strategy that assists in broadening the scope of existing approaches from a limited range of categories to a broader vocabulary. Moreover, we conduct an analysis of the domain gap between our dataset and real-world dataset, emphasizing the benefits of their combined use.

Our dataset will be publicly available to the research community, which will foster future research on more practical and robust 6D pose estimation algorithms and pave the way for broader applications.
\begin{table}[t]
  \caption{\textbf{Comparisons between Omni6D(-xl) and existing datasets.} Omni6D significantly extends the range of everyday object categories and instances.}
  \label{tab:dataset}
  \centering
  \scriptsize
  \begin{tabular}{@{}l|cc|ccccc@{}}
    \toprule
    Datasets & Mode & Realism & \# Categories & \# Instances & \# Images \\
    \midrule
    ShapeNet-SRN Cars~\cite{inerf} &  RGB & Synthetic & 1 & 3514 & - \\ 
        Sim2Real Cars~\cite{inerf} &  RGB & Real & 1 & 10 & - \\
        \midrule
        CAMERA~\cite{NOCS} & RGBD & Synthetic & 6 & 1085 & 0.3M \\
        REAL~\cite{NOCS} & RGBD & Real & 6 & 42  & 8k \\
        Wild6D~\cite{reponet} & RGBD & Real & 5 & 1722 & 1M \\
        \textbf{Omni6D} & RGBD & Real-Scanned & \textbf{166} & \textbf{4,688} &0.8M \\
        \textbf{Omni6D-xl} & RGBD & Real-Scanned & \textbf{419} & \textbf{15,957} &1.1M \\
        
    \bottomrule
  \end{tabular}
\end{table}
\section{Related Work}
\label{sec:related}

Existing work on category-level 6D object pose estimation can be generally divided into two types. After extracting features from images or point clouds, they compute Rotation, Translation, and Size (RTS) either through implicit point correspondence or explicit regression.

\noindent\textbf{Existing Datasets.} The most commonly used dataset for category-level 6D object pose estimation is NOCS~\cite{NOCS}, comprising both the synthetic CAMERA dataset and the real-world REAL dataset. CAMERA includes 300k RGBD images of 31 indoor scenes with 1,085 object instances across 6 categories, while REAL mirrors the categories in CAMERA and includes 8k RGBD images capturing 42 instances in 18 real scenes. Wild6D~\cite{reponet} consists of 5,166 videos with 1.1 million images over 1,722 object instances in 5 categories. ShapeNet-SRN Cars dataset and Sim2Real Cars dataset proposed in iNerf~\cite{inerf} both exclusively include a single car category. The former includes 3,514 instances derived from ShapeNet cars, while the latter is extracted from videos capturing 10 distinct unseen car models. These datasets are limited by their narrow range of categories, hindering their ability to generalize broadly.  Additionally, most training images are synthetic and lack realism, and their scenes are overly simplified, failing to account for common real-world challenges like occlusions.

\noindent\textbf{Implicit Methods. }
Implicit methods are based on point correspondence~\cite{NOCS, DualPoseNet, RBP, reponet, SGPA, CATRE, SPD, 6-PACK} . NOCS~\cite{NOCS}, one of the pioneering works in this area,  introduced the concept of Normalized Object Coordinate Space (NOCS). The final pose and size of the object are obtained by matching the predicted NOCS map with the observed depth input using the Umeyama algorithm~\cite{Umeyama} and RANSAC algorithm~\cite{RANSAC}. 

Subsequent algorithms such as DualPoseNet, RBP-Net and RePoNet~\cite{DualPoseNet, RBP, reponet} have continued to develop along the vein of NOCS, implicitly solving for pose after predicting the NOCS map. SPD~\cite{SPD} proposed a category-level shape prior, subsequently deforming this shape prior (i.e., average shape) to fit observed point cloud. SGPA, RePoNet, and CATRE~\cite{SGPA, reponet, CATRE} continue to develop along SPD's category-level shape prior approach. Algorithms like 6-PACK and SGPA~\cite{6-PACK, SGPA} extract low-rank structure points, i.e., keypoints, from dense observed point clouds. 6-PACK~\cite{6-PACK} predicts interframe motion of target instances through keypoint matching, while SGPA~\cite{SGPA} employs keypoints for more effective incorporation of sparse structural information during prior adaptation. These methods rely heavily on the RANSAC process to eliminate outliers, making them non-differentiable and time-consuming.

\noindent\textbf{Explicit Methods. }
Explicit methods are based on direct pose regression~\cite{DualPoseNet, RBP, CATRE, FS-Net,GPV, HS}. DualPoseNet and RBP-Net~\cite{DualPoseNet, RBP} conduct both explicit and implicit training,, where one parallel pose decoder explicitly regresses the pose. CATRE~\cite{CATRE}, recognizes the inherent difference between estimations of rotation and translation/size, explicitly regressing their residuals and carrying out an iterative pose estimation process. FS-Net~\cite{FS-Net} designs an autoencoder with 3D Graphic Convolution for latent feature extraction and separates the predictions for rotation and translation/size into two distinct networks: one estimates translation/size through two residuals, while the other handles rotation prediction by estimating deflections on two orthogonal axes. GPV-Pose and HS-Pose~\cite{GPV, HS} utilize the same foundational mechanism introduced by FS-Net~\cite{FS-Net}. GPV-Pose~\cite{GPV} proposes a decoupled confidence-driven rotation representation that facilitates geometrically-aware recovery of correlated rotation matrices and introduces a new geometry-guided point-by-point voting paradigm for robust retrieval of 3D object bounding boxes. Meanwhile, HS-Pose~\cite{HS} extends 3D-GC to extract mixed-range latent features from point cloud data through a simple network structure known as the HS layer.

\section{Omni6D Dataset}
\label{sec:dataset}

\subsection{Construction}

\noindent\textbf{Dataset Collection.} As shown in~\cref{tab:dataset}, \textbf{Omni6D} comprises 4,688 instances across an impressive span of 166 categories. Each instance is a high-resolution textured mesh, obtained using Shining 3D scanner\footnote{https://www.einscan.com/} and Artec Eva 3D scanner\footnote{https://www.artec3d.cn/}, collected from OmniObject3D~\cite{oo3d}. We normalize object models to fit within a $(-1,1)^3(m^3)$ three-dimensional space, and align objects within each category to a consistent canonical pose. In the latest dataset, \textbf{Omni6D-xl} builds upon and extends Omni6D, comprising 15,957 instances across an impressive span of 419 categories. For more details, please refer to Appendix Section C.

\noindent\textbf{Rendering.} We employ stratified sampling to split instances within each category, subsequently dividing them into training, validation, and test sets in a 7:2:1 ratio. In the construction of our dataset, we utilize 9 room models from the Replica dataset as backdrops. For each scenery setup, we randomly select a room model to act as the background, along with $6-8$ object instance models, which are allowed to perform free-fall motion within the room model, resulting in random scattering in a specific section of the room. Each object model is scaled by a random factor ranging from 0.8 to 1.2 as part of our data augmentation strategy. Considering the attention center of the combined instance models as the origin point, the camera randomly selects ten positions within a radius of $8-9~m$ and an elevation angle range between $30-90^\circ$. The camera then performs rendering at these selected positions while facing towards the attention center. 

\noindent\textbf{Setting.} We utilize BlenderProc 2.5.0~\cite{blenderproc} to implement the aforementioned rendering process. The intrinsic parameters of the camera are set to [577.5, 577.5, 319.5, 239.5], with an image size specified as $640\times480$. Our approach ensures the diversity and breadth of the dataset, making it suitable for rigorous testing and yielding accurate results.
\begin{figure*}[t]
  \centering
    \includegraphics[width=1\linewidth]{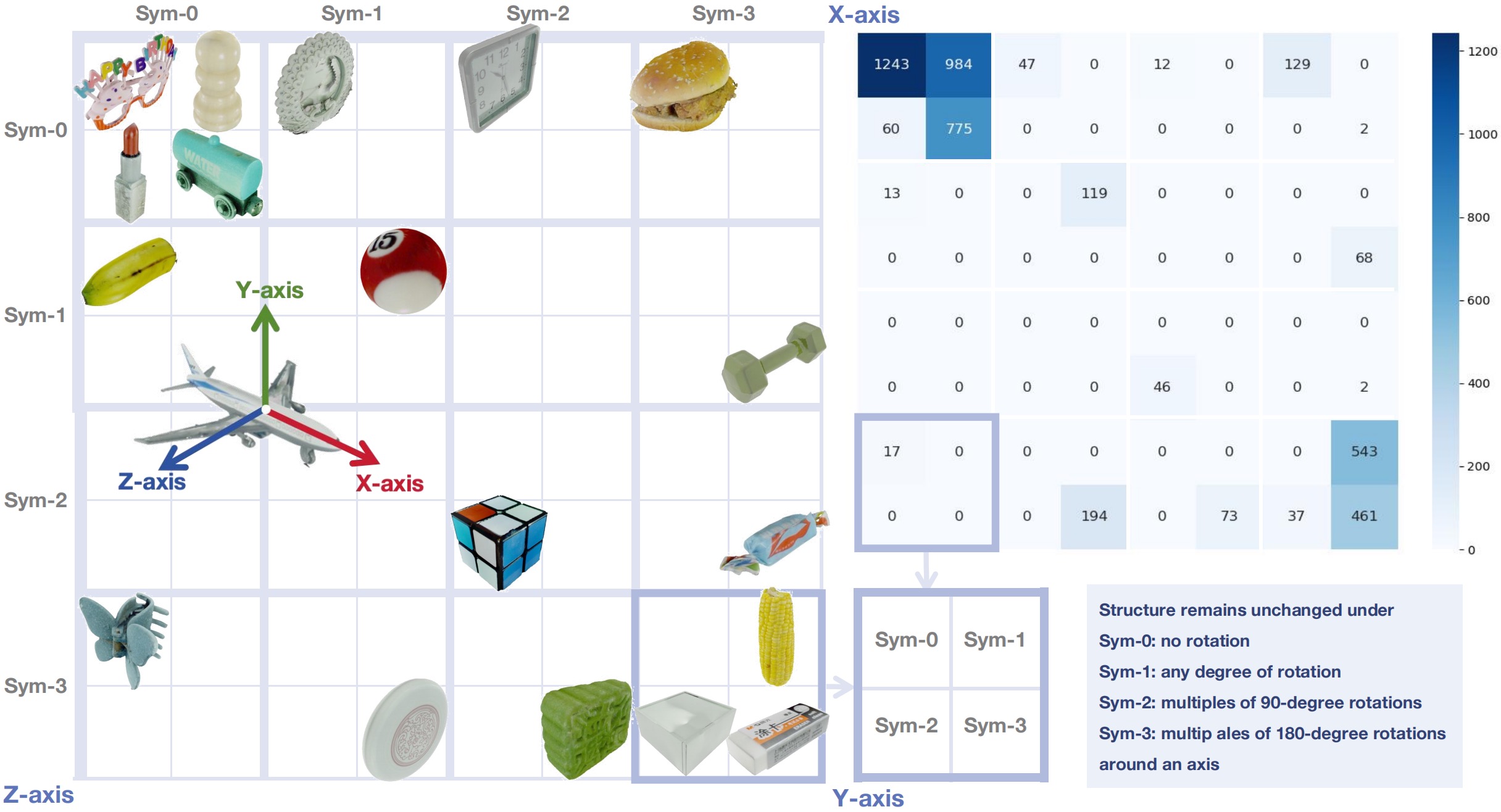}
   \caption{\textbf{Symmetry statistics.} The figure demonstrates different symmetry cases using object instances and provides a quantitative representation of the occurrence frequency for various combinations of distinct symmetry cases across the xyz-axes. }
   \label{fig:sym_info}
\end{figure*}
\begin{figure}[t]
  \begin{minipage}{0.55\textwidth}
      \centering
      \begin{subfigure}{0.48\linewidth}
        \centering
        \includegraphics[width=\linewidth]{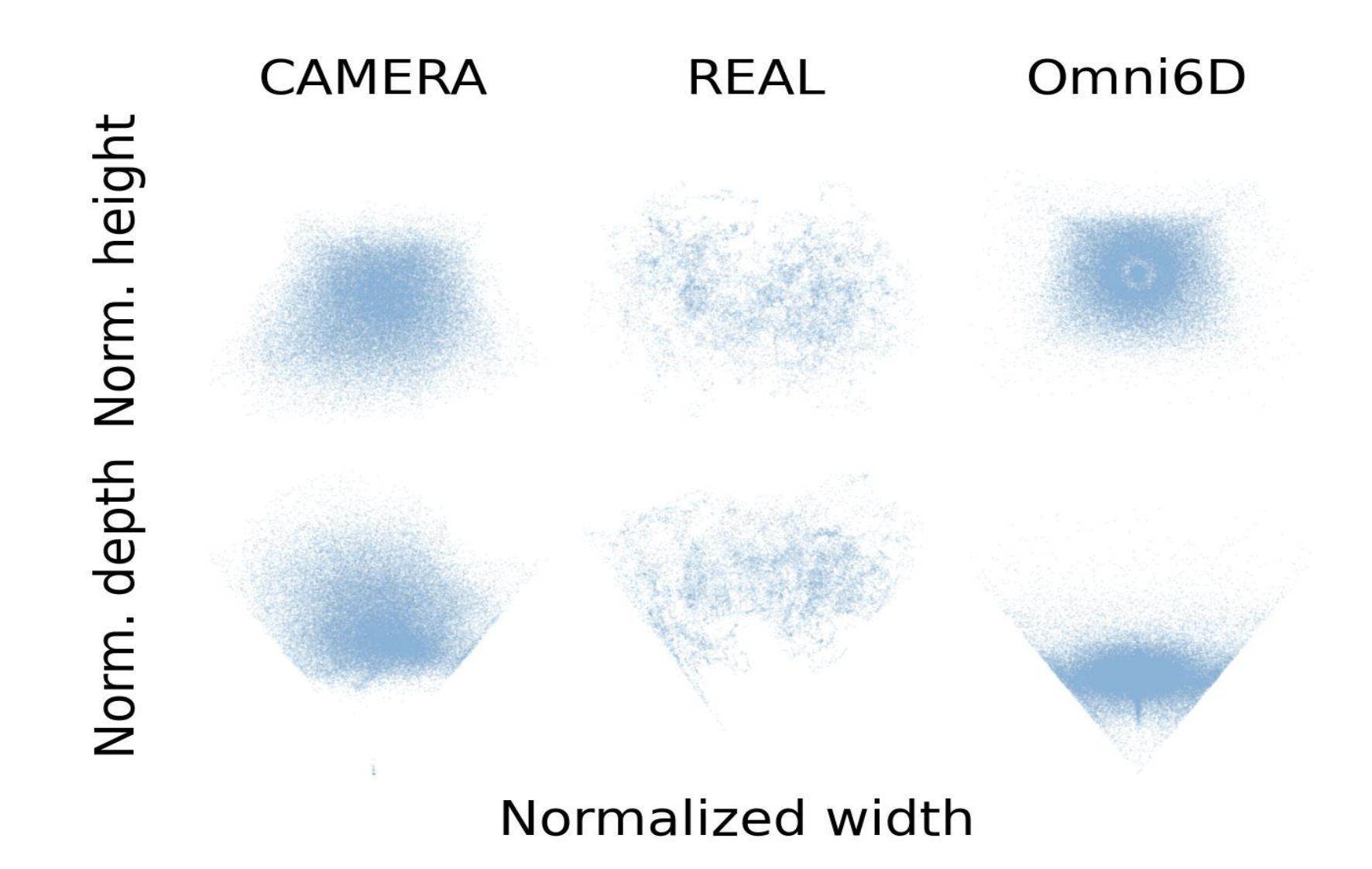}
        \caption{Point cloud centroids}
        \label{fig:data-a}
      \end{subfigure}
      \begin{subfigure}{0.48\linewidth}
        \centering
        \includegraphics[width=\linewidth]{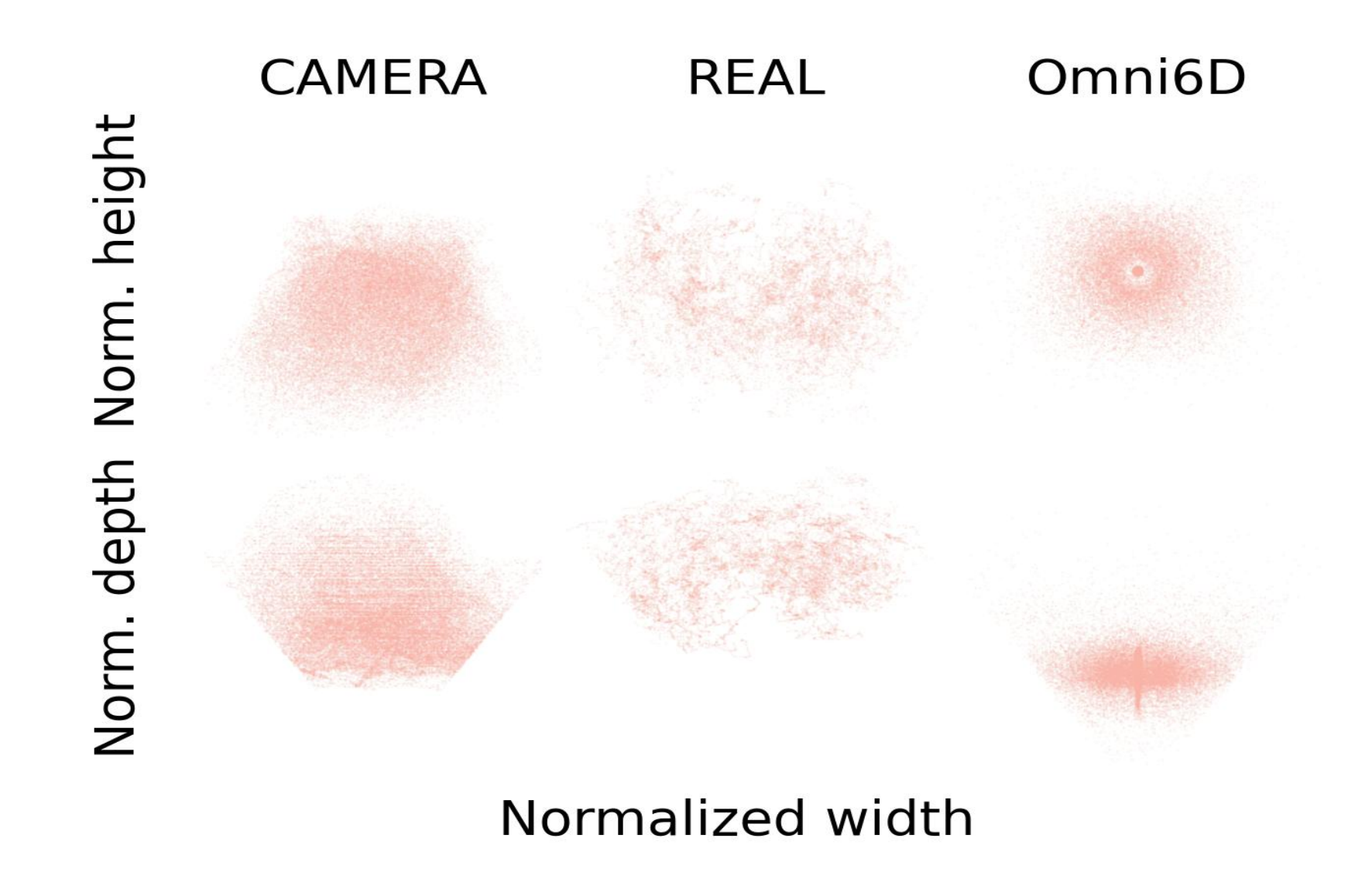}
        \caption{Object centroids}
        \label{fig:data-b}
      \end{subfigure}
      \begin{subfigure}{0.48\linewidth}
        \centering
        \includegraphics[width=\linewidth]{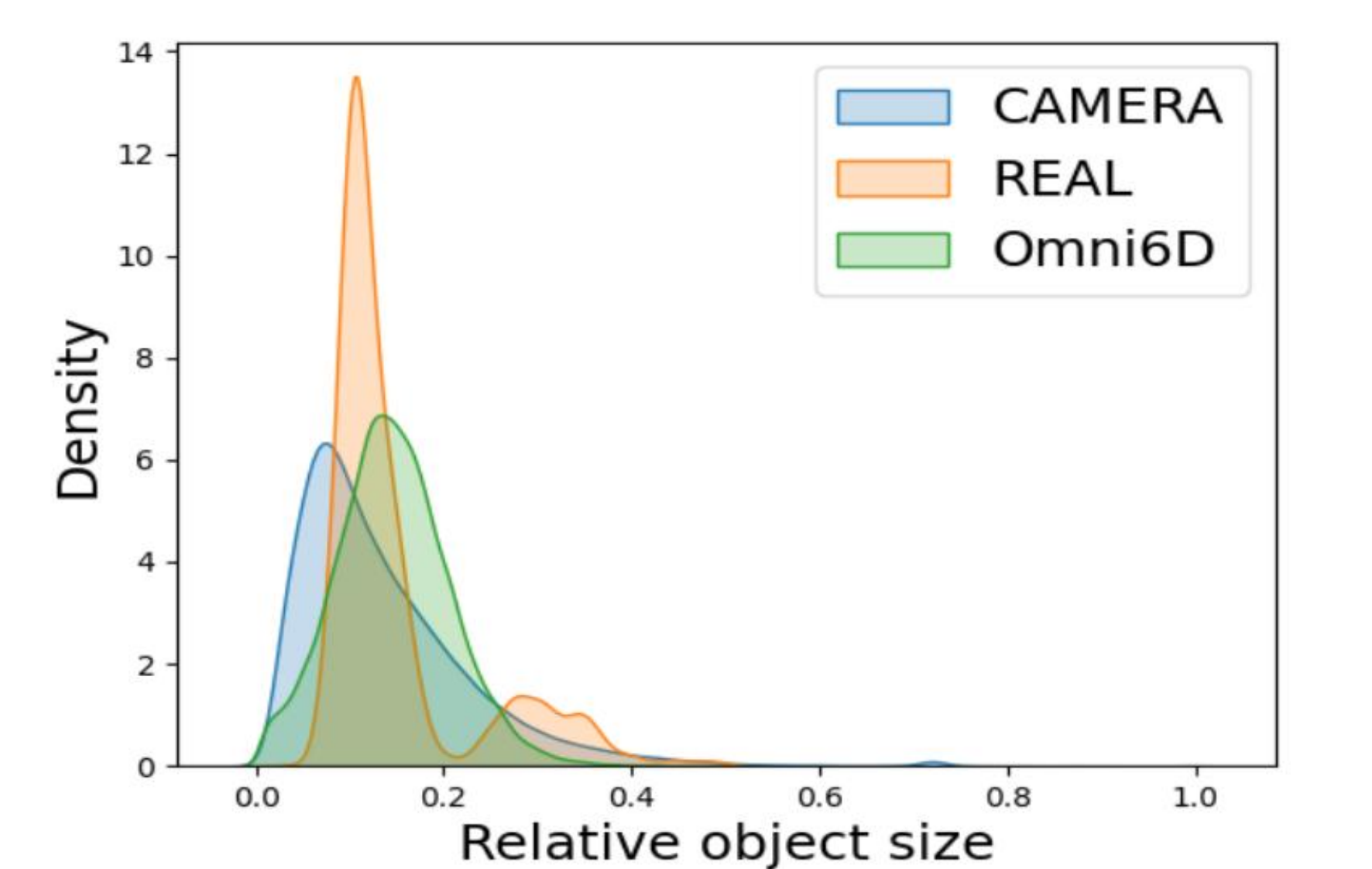}
            \caption{Relative 2D object size}
            \label{fig:data-c}
          \end{subfigure}
          \begin{subfigure}{0.48\linewidth}
            \centering
            \includegraphics[width=\linewidth]{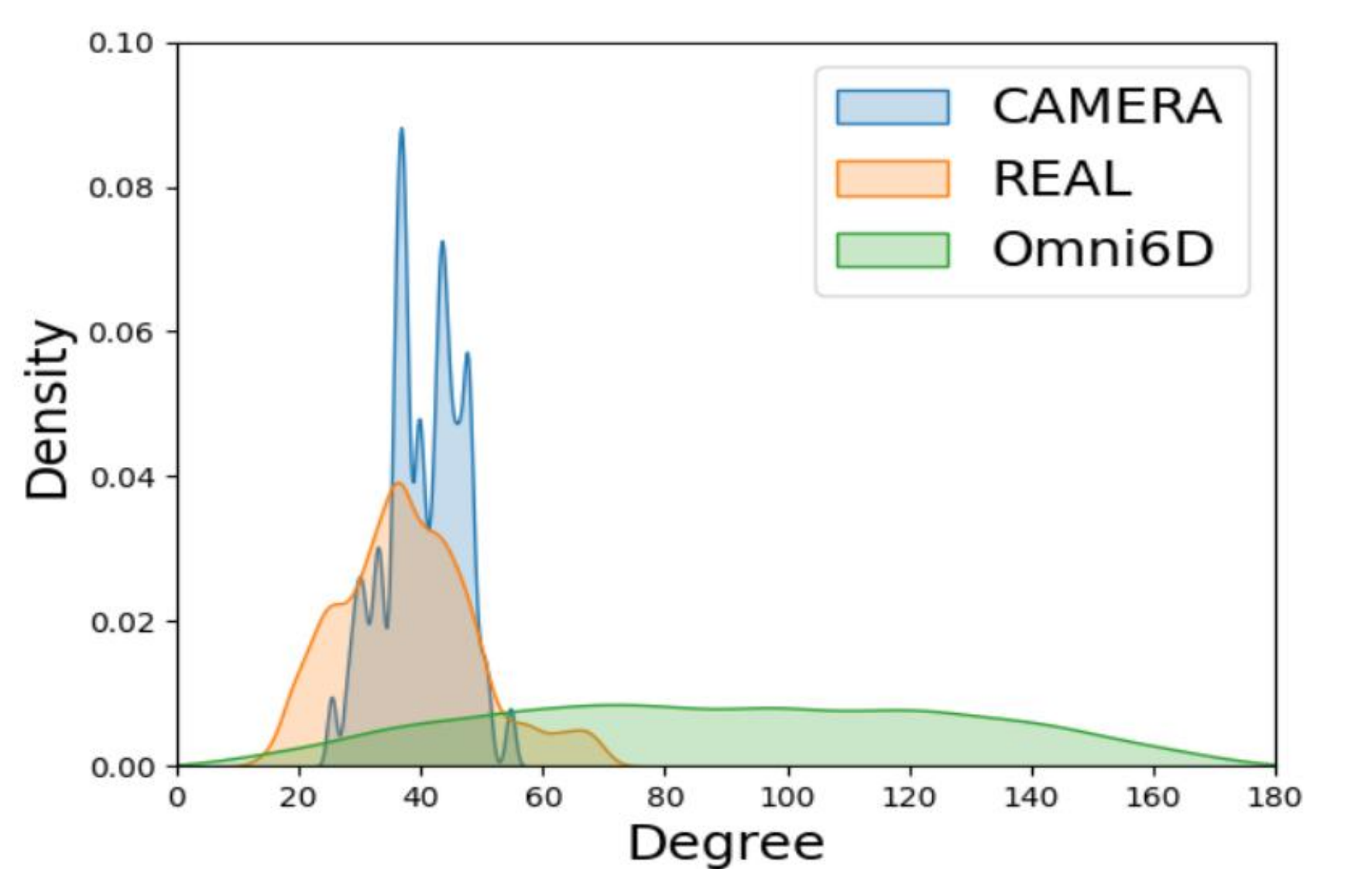}
            \caption{Angular deviation}
            \label{fig:data-d}
          \end{subfigure}
        \end{minipage}
        \begin{minipage}{0.36\textwidth}
          \begin{subfigure}{1\linewidth}
            \centering
            \includegraphics[width=\linewidth]{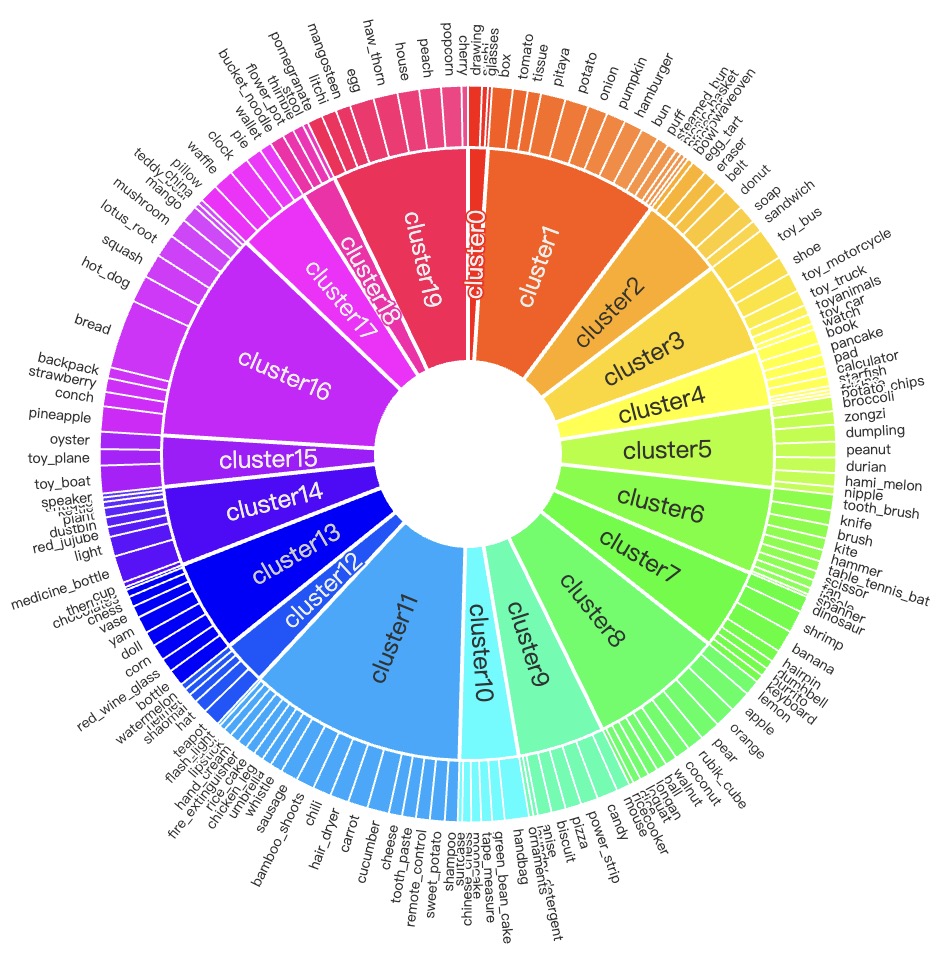}
            \caption{Clustering results}
            \label{fig:data-e}
          \end{subfigure}
        \end{minipage}
  \caption{\textbf{Omni6D analysis.} \textbf{(a)} distribution of point cloud centroids, \textbf{(b)} distribution of object centroids on (top) normalized image, XY-plane, and (bottom) normalized depth, XZ-plane, \textbf{(c)} density of relative 2D object size, \textbf{(d)} density of angular deviation from the upward direction, \textbf{(e)} Omni6D dataset clustering results. The angle of each sector in the chart reflects the relative size of the instance count within that category.}
  \label{fig:data_analysis}
\end{figure}

\subsection{Data Annotations}
\label{sec:annotation}
\noindent\textbf{Rich Annotations.} 
Each rendered output includes a rendered RGB image, instance mask, NOCS mapping~\cite{NOCS}, depth map, ground truth class label, as well as 6D pose and size. \cref{fig:teaser} exhibits a selection of rendered outputs. To reduce the storage size of the dataset, we encode high-precision depth maps into RGB images by multiplying depth by 10,000, rounding to nearest integer, and converting to base 256. The resulting three digits represent RGB channels.

\noindent\textbf{Rotational Invariance.} 
Rotational invariance implies that a symmetric object can retain its original shape after rotation by certain angles. Many common objects have this property. As shown in~\cref{fig:sym_info}, we define the coordinate system as a right-handed system with the x-axis pointing outwards and the y-axis oriented upwards. We contemplate three cases of rotational invariance where an object maintains its original shape after following rotations: any degrees (\textbf{Sym-1}), multiples of 90 degrees (\textbf{Sym-2}) and 180 degrees (\textbf{Sym-3}). Additionally, we denote the case of no rotational invariance around the axis as \textbf{Sym-0}. According to these definitions, all objects in Omni6D are annotated for their rotational symmetry around the xyz-axes. It's worth noting that symmetry attributes may differ among instances within the same category, requiring instance-level rather than category-level annotations. \cref{fig:sym_info} illustrates all kinds of symmetry cases using object instances and quantifies their occurrence frequency. \cref{fig:teaser} selects several examples to provide a more visual explanation of rotational invariance. These considerations are then integrated into our evaluation protocols in \cref{sec:sym}.

\subsection{Dataset Statistics}
\begin{figure}[t]
   \centering
      \begin{subfigure}{0.45\linewidth}
        \centering
        \includegraphics[width=\linewidth]{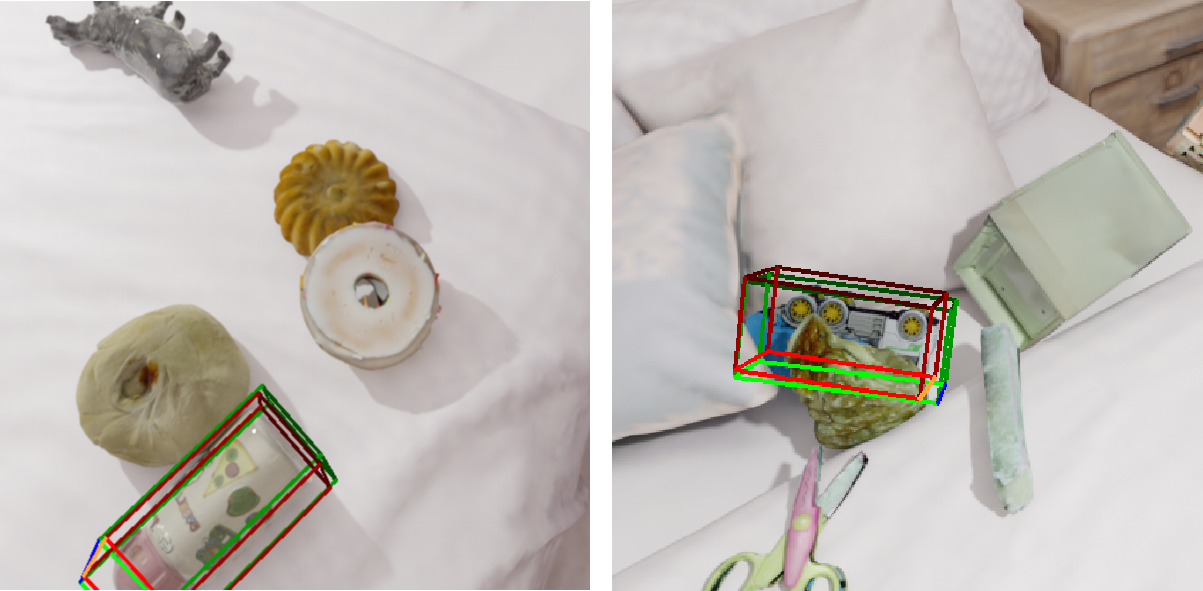}
        \caption{Challenges from occluded object}
        \label{fig:occlusion}
      \end{subfigure}
      \begin{subfigure}{0.45\linewidth}
        \centering
        \hfill
        \includegraphics[width=\linewidth]{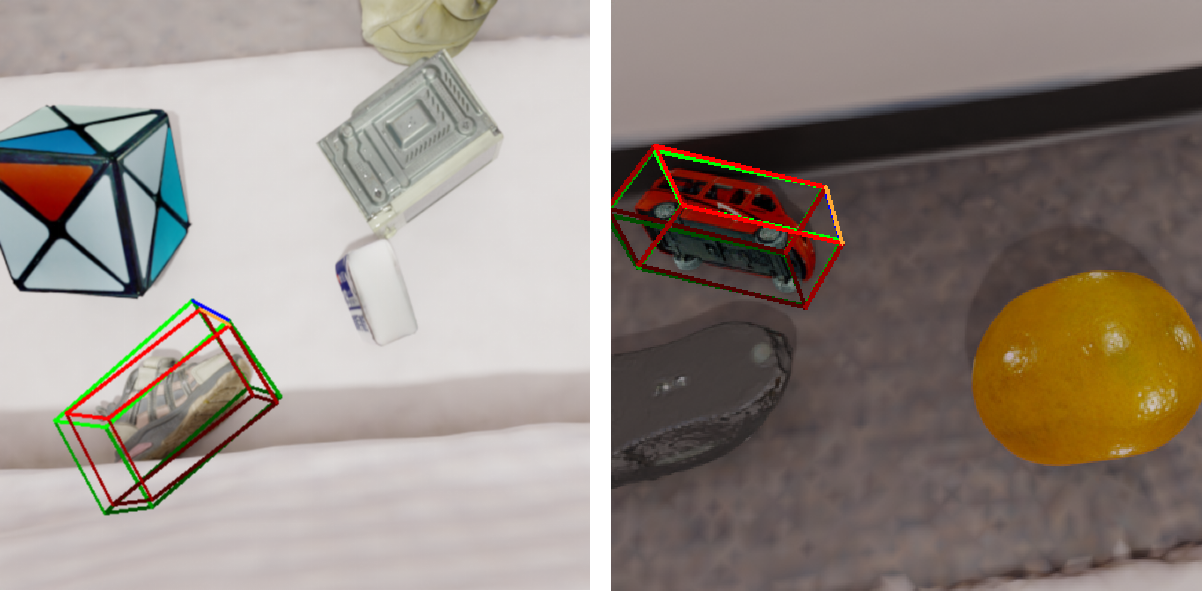}
        \caption{Challenges from bottom views}
        \label{fig:lower_bottom}
      \end{subfigure}
   \caption{\textbf{Challenges from Omni6D.} \textbf{(a)} Algorithms trained on Omni6D can overcome challenges in estimating poses for occluded object instances. The \textbf{left} shows an occluded object instance at the edge of the image, while the \textbf{right} image shows an object instance obstructed by other objects. \textbf{(b)} Algorithms trained on Omni6D can accurately estimate poses with only the lower half or bottom appearance of an object. The green and red colors respectively denote the ground truth and predicted 3D bounding boxes. The blue and orange lines on the boxes separately highlight the intersecting lines of the frontal face and the top face of the two 3D bounding boxes, while the darker lines indicate the bottom of the bounding boxes.}
   \label{fig:challenge}
\end{figure}

\noindent\textbf{Spatial Statistics.}
Omni6D aims to overcome challenges in estimating poses for occluded object instances. \cref{fig:data-a} and \cref{fig:data-b} show the spatial distribution of point clouds and objects by projecting their centroids on the XY-plane (top) and XZ-plane (bottom)~\cite{Omni3D}. \cref{fig:data-c} depicts the relative object size distribution, defined as the square root of the object-to-image area ratio. We observe that the spatial distribution of Omni6D is similar to that of CAMERA and REAL, with a greater resemblance to CAMERA despite having a closer depth range. However, a more pronounced discrepancy between the spatial distribution of point clouds and objects is evident in Omni6D compared to CAMERA and REAL. This observation suggests a higher occurrence of occlusion scenes in Omni6D, highlighting the intricate challenges it presents to 6D object pose estimation.  Nonetheless, as depicted in~\cref{fig:occlusion}, algorithms trained on Omni6D demonstrate their robustness in tackling these complexities.

\noindent\textbf{Angular Deviation.}
Omni6D enables accurate pose estimation using only the lower half or bottom appearance of objects. \cref{fig:data-d} depicts the density of angular deviations from the upward direction, \ie y-axis. Our dataset displays a more uniform distribution of object angles relative to the upward axis and exhibits greater deviation from the canonical pose angles. 
Unlike NOCS, which primarily uses upright object placement, Omni6D utilizes physical simulations for free-fall object positioning~\cite{blenderproc}. As a result, it presents more challenging and diverse pose estimation scenes. Training on Omni6D enhances algorithms' robustness to object rotation angles, as evidenced by the image in \cref{fig:lower_bottom}.

\noindent\textbf{Shape Priors.} We obtain the mean latent embedding and shape prior for each category from the variational autoencoder~\cite{Shape_prior}. \cref{fig:teaser} showcases categorical shape priors, each displaying unique characteristics, facilitating an intuitive association between point cloud shapes and corresponding real-world entities. 
Meanwhile, \cref{fig:data-e} explains clustering results based on categorical latent embeddings, where we employ agglomerative clustering~\cite{cluster} to group categories into 20 clusters. It highlights the geometric coherence among semantically identical objects (especially man-made ones) in Omni6D dataset and further confirms that these categorical shape priors can effectively leverage the wealth of shape information from numerous similar objects to elucidate category features. These insights provide a theoretical basis for applications of category-level 6D object pose estimation using our Omni6D dataset. 

\section{Evaluation and Analysis}
\label{sec:experiments}
\subsection{Experimental Setup}
\label{sec:setup}

\noindent\textbf{Datasets.}
Our experimentation utilized two datasets, namely Omni6D and Omni6D$_{out}$. Omni6D are partitioned into training, validation, and test sets in a 7:2:1 ratio, denoted as Omni6D$_{train}$, Omni6D$_{val}$ and Omni6D$_{test}$ respectively. These sets are further subdivided into subsets with increasing category sizes of 3, 6, 12, 24, and 48. We denote the subset containing $n$ categories as cls$n$. Each subset includes all classes present in the previous subset with additional classes included to meet the desired total. \cref{fig:finetune_cls} presents the specific categories included in cls$n$ and their respective sizes relative to each other. Omni6D$_{out}$ is utilized as an additional test set to measure our algorithm's inter-category generalization. This dataset, constructed similarly to Omni6D, encompasses 52 models spanning 17 categories unseen in Omni6D, along with 4762 images. For additional details on datasets, please refer to the appendix.

\noindent\textbf{Details.}
All experiments are carried out on a server equipped with an Intel(R) Xeon(R) Gold 6248R CPU @ 3.00GHz and an NVIDIA A100-SXM4-80GB GPU. We maintain consistency in parameters and strategies throughout training, ensuring uniformity in our experiment environment. Given the challenges of semantic classification with a large vocabulary, we use ground truth masks to mitigate the impact of low-quality classification on pose estimation results.

\subsection{Symmetry-Aware Evaluation}
\label{sec:sym}

\noindent\textbf{Basic Evaluation Metrics.}
We utilize the average accuracy of Intersection over 3D Union (IoU)~\cite{KITTI} in object detection, and $n^\circ m~cm$ in pose estimation. We further decompose $n^\circ m~cm$~\cite{rot1, DeepIM} to individually evaluate the model's predictive error $n^\circ$ for pose and $m~cm$ for translation. For these three types of errors, the thresholds considered are $\{50\%, 75\%\}$, $\{5^\circ, 10^\circ\}$ and $\{2~cm, 5~cm\}$~\cite{PoseCNN, Uncertainty, BB8}. Additionally, we set a detection threshold for objects requiring at least a 10\% overlap between predicted and ground-truth bounding boxes.

\noindent\textbf{Our Symmetry-Aware Metrics.} 
Due to NOCS's limited categories, traditional algorithms mainly handle basic symmetry cases, such as rotational symmetry around the y-axis. However, Omni6D has a wider range of objects with different rotational invariances across multiple axes. \cref{fig:sym_info} provides symmetry statistics for Omni6D objects. To alleviate this issue, we propose a symmetry-aware metric. Unlike prior works focusing solely on the y-axis, our method considers rotation symmetry around all three axes.

We define the relevant variables as follows: $L_s$ denotes our symmetry-aware metric, $L$ denotes the original metric. $R$ stands for the ground truth rotation matrix, while $R^*$ represents the predicted rotation matrix. $R^*_{\theta_x, \theta_y, \theta_z}$ corresponds to the predicted rotation matrix after sequentially rotating by $\theta_x$, $\theta_y$, and $\theta_z$ degrees around the xyz axes. The rotational invariance cases around the x, y, and z axes are denoted as Sym-$n_x$, Sym-$n_y$, and Sym-$n_z$, where $n_x$, $n_y$, and $n_z$ are the respective rotation parameters. Objects that align with Sym-$n$ around an axis maintain their original shape when rotated by an angle from $\Theta_n$.

\begin{algorithm}[t]
\scriptsize
\caption{Compute Our Symmetry-Aware Metric $L_s$}
\label{alg:my_algorithm}
\begin{algorithmic}[1]
\Procedure{symmetric\_metric}{$L$, $R$, $n_x$, $n_y$, $n_z$}
\State $\Theta_0 = \{0^\circ\}$
\State $\Theta_2 = \{0^\circ, 90^\circ, 180^\circ , 270^\circ\}$
\State $\Theta_3=\{0^\circ, 180^\circ\}$  \hfill  \emph{// Rotations around Sym-1 axis need not be considered.}
\State $c = \text{count}( \text{1 occurrences in } \{\text{$n_x$, $n_y$, $n_z$}\})$
\If {$c \geq 2$} \hfill \emph{// The object is a sphere.}
\State $L_s = L(R^*, R)$
\ElsIf {$c == 1$}\hfill \emph{// Rotations around Sym-1 axis can be disregarded.}
\State Without loss of generality, assume $n_x==1$.
\State $L_s = \min_{\theta_y\in\Theta_{n_y}, \theta_z\in\Theta_{n_z}}L(R^*_{\theta_y, \theta_z}, R)$ 
\ElsIf {$c == 0$}\hfill \emph{// Simply enumerate all cases.}
\State $L_s = \min_{\theta_x\in\Theta_{n_x}, \theta_y\in\Theta_{n_y}, \theta_z\in\Theta_{n_z}}L(R^*_{\theta_x, \theta_y, \theta_z}, R)$
\EndIf
\State \Return $L_s$
\EndProcedure
\end{algorithmic}
\end{algorithm}


Since the Euler angles are compact~\cite{slam}, the most straightforward approach is to determine the category of rotational invariance for each axis \{x, y, z\} sequentially, as mentioned in \ref{sec:annotation}. To simplify computations, we set $\Theta_0 = \{0^\circ\}$, $\Theta_1 = \{0^\circ, 1^\circ, ... , 359^\circ\}$, $\Theta_2 = \{0^\circ, 90^\circ, 180^\circ , 270^\circ\}$, $\Theta_3=\{0^\circ, 180^\circ\}$. We can define $L_s$ as $L_s = \min_{\theta_x\in\Theta_{n_x}, \theta_y\in\Theta_{n_y}, \theta_z\in\Theta_{n_z}}L(R^*_{\theta_x, \theta_y, \theta_z}, R)$.

However, due to the singularity of Euler angles~\cite{slam}, we can simplify the above rotation transformation. The pseudo-code implementation of our Symmetry-Aware Evaluation is provided
in Algorithm~\ref{alg:my_algorithm}. It allows us to simplify what was originally at most $360^3$ computations to a maximum of only $4^3$ computations.


\subsection{Large-Vocabulary 6D Pose and Size Estimation}
\label{sec:experiment}
\noindent\textbf{Performance on Omni6D.}
We present results of algorithms~\cite{SPD,SGPA,RBP,GPV,HS} trained on Omni6D$_{train}$ and tested on Omni6D$_{test}$. We compare their quantitative results in \cref{tab:cls166_omni6d} and their qualitative results in Fig. S10 in Appendix. 
Additionally, we compare the quantitative results of algorithms trained on Omni6D-xl$_{train}$ and tested on Omni6D-xl$_{test}$ in \cref{tab:cls339_omni6dxl}. 
The performance disparity among algorithms for category-level 6D object pose estimation becomes markedly pronounced when applied to large-vocabulary datasets, in contrast to the more consistent performance previously observed on the Real and CAMERA datasets\cite{NOCS}. This highlights the inherent strengths and weaknesses across various model structures.

This observation suggests the potential importance of our large-vocabulary dataset in uncovering the relative performance of different models. It appears that the increased complexity of the dataset could push model architectures to their theoretical limits, possibly revealing intrinsic characteristics otherwise obscured in less complex scenarios. For example, SPD, SGPA is particularly proficient in predicting rotation, and SPD achieves the highest score in $n^\circ m~cm$. This could be due to its implicit network's propensity for generating more reliable rotational forecasts. Meanwhile, DualPoseNet and HS-Pose provide more accurate predictions for translation and score higher in IoU. This could be associated with the characteristic of models with explicit networks to produce better translations and size estimates.
 
Our large-vocabulary dataset, encompassing a broad spectrum of shapes and appearances, enables a comprehensive evaluation of diverse category-level pose estimation methods. This serves not only as a robust test of an algorithm's generalizability but also as a valuable tool in understanding the advantages offered by different algorithmic structures.

\begin{table*}[t]
    \caption{\textbf{Category-level performance on Omni6D dataset.} Models are trained on Omni6D$_{train}$ and tested on Omni6D$_{test}$. Instances within each category in the test set are unseen during training, substantiating the algorithms' capacity to generalize within individual categories under large-vocabulary settings. \textbf{Bold} and \underline{underlined} results indicate the best and second-best performers.}
    \label{tab:cls166_omni6d}
    \centering
    \scriptsize
    \begin{tabular}{@{}lc|cc|cccc|cccc@{}}
        \toprule
        Methods & Network & $IoU_{50}$ & $IoU_{75}$ & $5^\circ2cm$ & $5^\circ5cm$ & $10^\circ2cm$ & $10^\circ5cm$ & $5^\circ$ & $10^\circ$  & $2cm$  & $5cm$\\ 
        \midrule
        SPD~\cite{SPD} & implicit & 44.56 & 20.37 & \underline{7.55} & \textbf{9.56} & \underline{14.76} & \textbf{19.23} & \textbf{10.68} & \textbf{21.02} & 37.49 & 70.09\\ 
        SGPA~\cite{SGPA} & implicit & 36.34 & 14.44 & 4.78 & 6.84 & 10.13 & 15.03 &8.49 & 17.73 & 25.57 & 59.18\\ 
        DualPoseNet~\cite{DualPoseNet} & hybrid & \underline{58.84}	&\textbf{25.49}&	\textbf{8.28}	&\underline{9.30}	&\textbf{17.26}	&\underline{19.05}	&\underline{9.38} & \underline{19.18} & \underline{73.82} & \underline{96.37}\\ 
        RBP-Pose~\cite{RBP} & hybrid & 35.92 & 4.66 & 0.37 & 0.60 & 0.53 & 0.80 &0.75 & 0.96 & 39.73 & 83.55\\ 
        GPV-Pose~\cite{GPV} & explicit & 15.28 & 0.26 & 0.10 & 0.70 & 0.14 & 0.96 & 2.25 & 2.96 & 5.31 & 33.70 \\ 
        HS-Pose~\cite{HS} & explicit & \textbf{62.65} & \underline{23.02} & 4.26 & 4.85 & 10.49 & 11.61 & 4.96 & 11.75 & \textbf{80.93 }& \textbf{97.78} \\  
        \bottomrule
    \end{tabular}
\end{table*}

\begin{table*}[t]
    \caption{\textbf{Category-level performance on Omni6D-xl dataset.} Models are trained on Omni6D-xl$_{train}$ and tested on Omni6D-xl$_{test}$.}
    \label{tab:cls339_omni6dxl}
    \centering
    \scriptsize
    \begin{tabular}{@{}lc|cc|cccc|cccc@{}}
        \toprule
        Methods & Network & $IoU_{50}$ & $IoU_{75}$ & $5^\circ2cm$ & $5^\circ5cm$ & $10^\circ2cm$ & $10^\circ5cm$ & $5^\circ$ & $10^\circ$  & $2cm$  & $5cm$\\ 
        \midrule
        SPD~\cite{SPD} & implicit & 42.28 & 16.73 & 3.42 & \underline{6.47} & 6.45 & \underline{12.16} & \textbf{10.62} & \underline{17.71} & 21.05 & 54.14\\ 
        SGPA~\cite{SGPA} & implicit & 37.62 & 12.48 & 2.45 & 5.56 & 5.12 & 10.98 & \underline{10.25} & \textbf{17.88} & 16.21 & 47.31\\ 
        DualPoseNet~\cite{DualPoseNet} & hybrid & \underline{59.15} & \underline{24.26} & \textbf{5.92} & \textbf{7.24} & \textbf{10.74} & \textbf{13.00} & 7.70 & 13.72 & \underline{65.05} & \underline{87.44}\\ 
        RBP-Pose~\cite{RBP} & hybrid & 41.21 & 6.55 & 0.52 & 1.63 & 0.59 & 1.74 & 3.02 & 3.20 & 23.20 & 57.27\\ 
        GPV-Pose~\cite{GPV} & explicit & 17.01 & 1.42 & 0.51 & 1.90 & 0.55 & 2.07 & 5.08 & 5.72 & 5.19 & 26.32 \\ 
        HS-Pose~\cite{HS} & explicit & \textbf{62.92} & \textbf{26.50} & \underline{4.98} & 6.26 & \underline{8.81} & 10.73 & 6.65 & 11.39 & \textbf{71.68} & \textbf{89.38} \\  
        \bottomrule
    \end{tabular}
\end{table*}

\begin{table}[t]
    \caption{\textbf{Category-level performance on unseen categories.} Models are trained on Omni6D$_{train}$ and tested on Omni6D$_{out}$. Categories in the test set never appear in the training set, validating the algorithms' ability to generalize across categories.     }
    \label{tab:cls166_unseen}
    \centering
    \scriptsize
    \begin{tabular}{@{}lc|cc|cccc|cccc@{}}
    \toprule
        Methods & Network & $IoU_{50}$ & $IoU_{75}$ & $5^\circ2cm$ & $5^\circ5cm$ & $10^\circ2cm$ & $10^\circ5cm$ & $5^\circ$ & $10^\circ$  & $2cm$  & $5cm$\\ 
        \midrule        
        SPD~\cite{SPD} & implicit & 7.56 & 0.95 & 0.18 & 0.40 & 0.80 & 1.65 & 0.65 & 2.36 & 8.88 & 40.59 \\ 
        SGPA~\cite{SGPA} & implicit & 7.05 & 0.60 & 0.07 & 0.28 & 0.19 & 0.82 & 0.53 & 1.69 & 3.87 & 28.28\\ 
        DualPoseNet~\cite{DualPoseNet} & hybrid &  \textbf{36.85} & \textbf{12.06} & \textbf{3.24 }& \textbf{3.37} & \textbf{8.04 }& \textbf{8.51} & \textbf{3.39 } & \textbf{8.64 } & \underline{78.00} & \underline{98.60 }\\ 
        RBP-Pose~\cite{RBP} & hybrid & 26.18 & 1.95 & 0.01 & 0.02 & 0.02 & 0.03 & 0.02 & 0.03 & 16.74 & 43.06 \\ 
        GPV-Pose~\cite{GPV} & explicit & 10.97 & 0.14 & 0.03 & 0.18 & 0.12 & 0.57 & 0.30 & 1.07 & 7.14 & 41.30 \\ 
        HS-Pose~\cite{HS} & explicit & \underline{36.75} & \underline{8.92} &\underline{ 1.54} & \underline{1.66} & \underline{4.67} &\underline{ 5.16} & \underline{1.75 }& \underline{5.38 }& \textbf{79.95 } & \textbf{98.27 }\\ 
        \bottomrule
    \end{tabular}
\end{table}

\begin{figure}[t]
  \centering
   \includegraphics[width=1\linewidth]{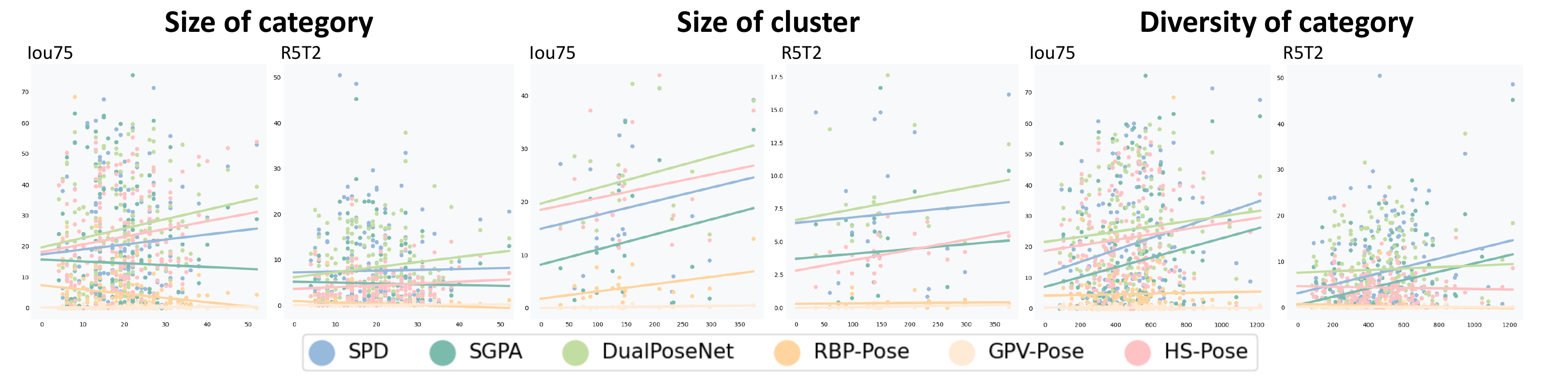}
   \caption{\textbf{Category-Wise Performance on Omni6D Dataset.} The x-axis, moving from left to right, sequentially represents: the number of objects within a category (\textbf{Semantic Category}), the number of objects within a cluster clustered based on shape priors (\textbf{Shape Category}) and the diversity of instances within a category. The y-axis depicts category or clustered group results for IoU$_{75}$ and $5^\circ2~cm$ metrics. Each plotted point illustrates the algorithm's result for a specific category or cluster, while the line showcases the trend of the linear fit for the scattered points.}
   \label{fig:cate}
\end{figure}
\begin{figure}[t]
  \begin{subfigure}{0.41\linewidth}
      \centering
    \includegraphics[width=\linewidth]{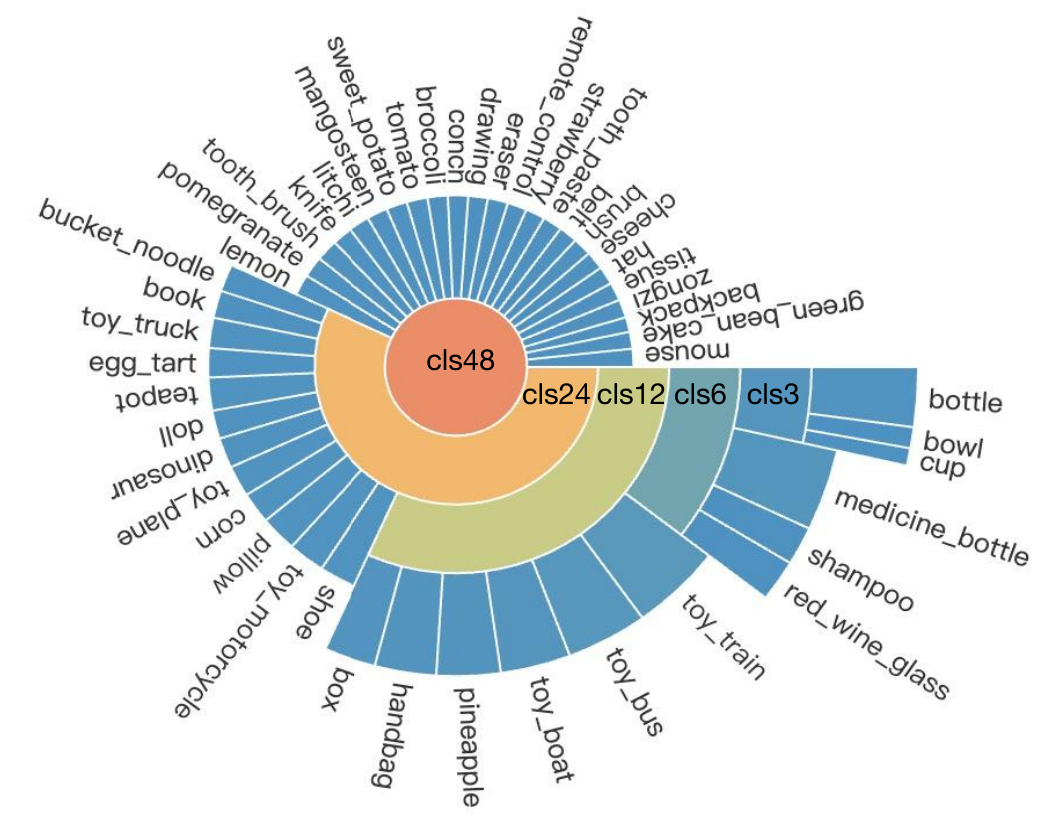}
       \caption{Category inventory of cls$n$}
       \label{fig:finetune_cls}
  \end{subfigure}
  \begin{subfigure}{0.55\linewidth}
      \centering       \includegraphics[width=\linewidth]{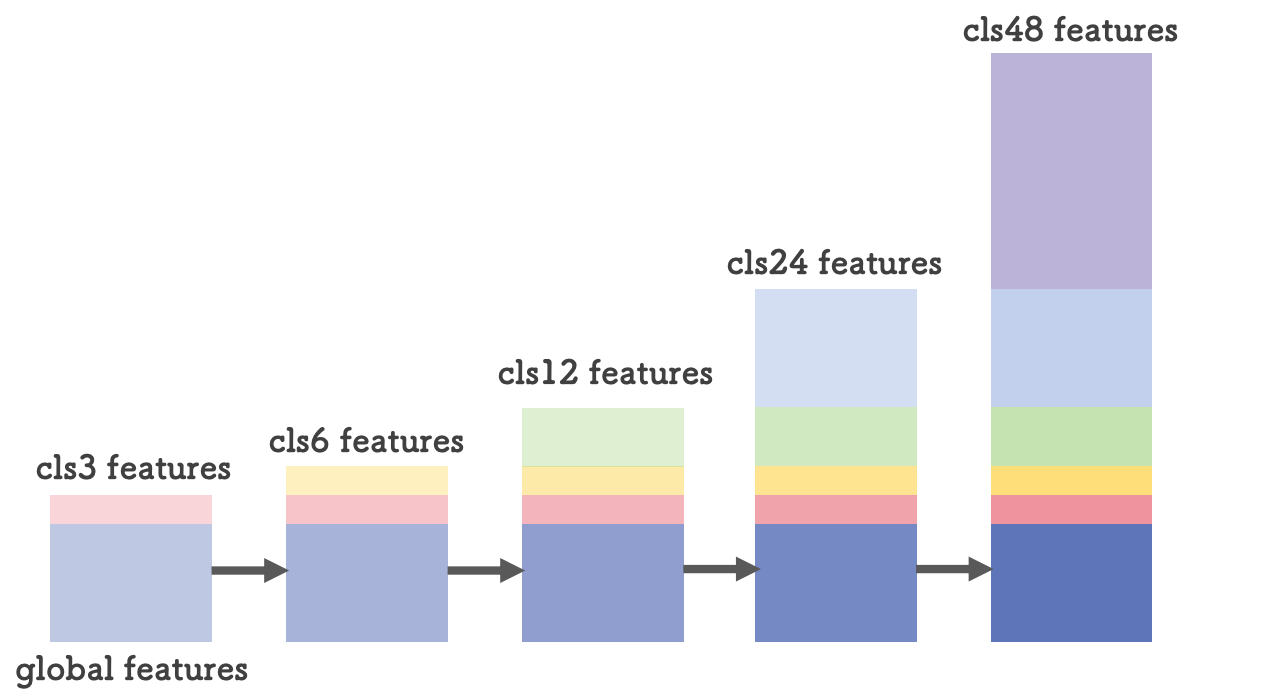}
       \caption{Our finetune strategy}
       \label{fig:strategy}
  \end{subfigure}
    \caption{\textbf{Our finetune strategy.} \textbf{(a)} Category inventory of cls$n$ within Omni6D dataset. The angle of each sector in the chart reflects the relative size of the instance count within that category. \textbf{(b)} In each fine-tuning step, we double the category count, copying trained global features and old category parameters into the new network while initializing the new category parameters. An observable deepening of color is indicative of the escalating count of training iterations.}
   \label{fig:sym_info}
\end{figure}
\begin{figure*}[t]
  \centering
   \includegraphics[width=0.9\linewidth]{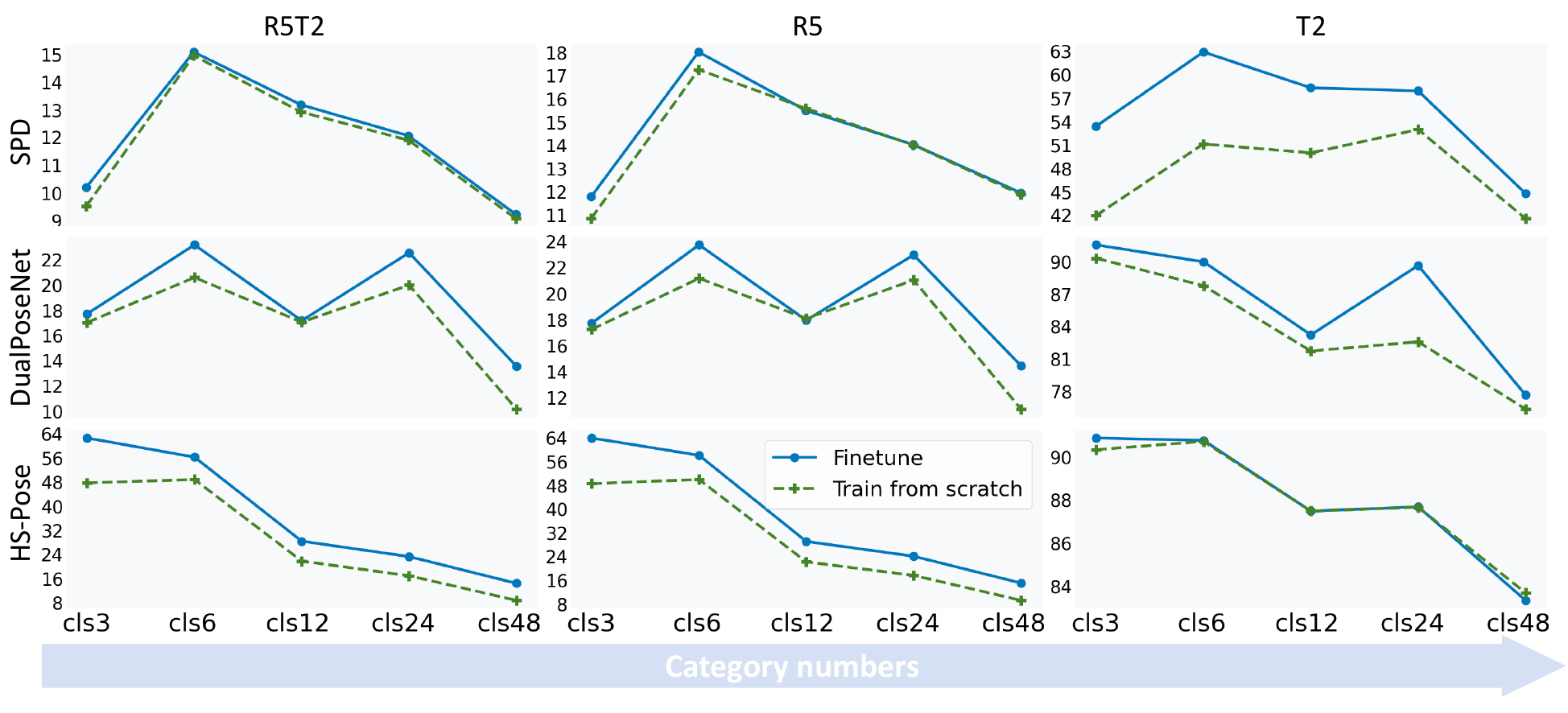}   \caption{\textbf{Finetuned results.} 
   Each figure's x-axis represents the number of categories in the training and test set, while the y-axis displays the outcomes of $5^\circ2~cm$, $5^\circ$ and $2~cm$ metrics. Each row, from top to bottom, sequentially employs three methods: SPD~\cite{SPD}, DualPoseNet~\cite{DualPoseNet}, and HS-Pose~\cite{HS}. The figures depict the outcomes derived from two training strategies as the number of training categories increases, accompanied by the gradual expansion of corresponding test sets.}
   \label{fig:finetune}
\end{figure*}

\noindent\textbf{Generalization Performance.} We evaluate algorithms on Omni6D$_{out}$ to assess their inter-category generalization capabilities. The outcomes are presented in ~\cref{tab:cls166_unseen}. Notably, DualPoseNet and HS-Pose emerged as superior performers, outclassing others across all metrics, thereby demonstrating excellent generalization abilities. Contrastingly, implicit methods including SPD and SPGA exhibited marked limitations. Qualitative results are shown in Fig. S11 in Appendix.

Drawing parallels with the observations from \cref{tab:cls166_omni6d}, we found that metrics such as translation and IoU were relatively easier to excel in, suggesting superior generalization abilities in translation and size prediction. Conversely, the generalization of rotation emerges as a considerable challenge in category-level 6D object pose estimation, especially within large-vocabulary scenes.

\noindent\textbf{Category-wise Analysis.}
Based on the IoU$_{75}$ and $5^\circ2~cm$ metrics, we conducted a detailed category-wise analysis of the results from~\cref{tab:cls166_omni6d}. Left columns in~\cref{fig:cate} illustrate the correlation between category-level 6D pose estimation performance and the number of instances within each category in Omni6D$_{train}$. 
Middle columns in ~\cref{fig:cate} analyze the correlation between cluster-level average performance and cluster size based on the clustering results described in ~\cref{fig:data-e}.
We found that the performance of pose estimation for each category is more strongly correlated with the number of instances within clusters than with semantic categories, showing a positive correlation. This suggests that shape categories have a greater impact on training than semantic categories do. Notably, algorithms like SPD, SGPA, and RBP-Pose that utilize shape prior structures are particularly sensitive to this influence.

Right columns in~\cref{fig:cate} reveal the correlation of pose estimation performance relative to instance diversity within each category in the training set. We measured instance diversity by calculating the mean chamfer distance~\cite{chamfer} among all pairs of instances in each category. The results show that as diversity within a category increases, pose estimation performance tends to improve. This observation aligns with the assertion made by~\cite{Prior-free}: The key to the success of prior-based methods lies in the deformation modules, which learns to synthesize world-space target objects and explicitly builds the correspondence between camera and world-space. As the number of instances increases and the diversity within a shape category expands, the model's capacity to learn deformation from priors to actual instance shapes is strengthened, leading to improved results.
\subsection{Fine-Tuning from Limited Categories}
\label{sec:finetune}
We propose a finetuning strategy that helps extend methods from a limited set of categories to large-vocabulary. We take SPD\cite{SPD}, DualPoseNet\cite{DualPoseNet}, and HS-Pose\cite{HS} as examples which belong to three different network architectures and show good performance on Omni6D$_{test}$. We respectively take their best models on CAMERA as our pre-trained models.

Initiating the fine-tuning process, we utilize three categories: bottle, bowl, and cup, which are concurrently present in both Omni6D and CAMERA datasets, aligning with the cls3 category. By facilitating the training on Omni6D-cls3, we enable a transfer of the model from CAMERA to Omni6D. Following the method illustrated in \cref{fig:strategy}, we engage in an iterative fine-tuning process on a progressively expanded category dataset until it reaches our desired number. In our experiments, we set this target number to be 48 categories. 

In parallel, we conduct training from scratch separately on cls3, cls6, ..., and cls48 as a comparison, employing the same number of training iterations. As shown in \cref{fig:finetune}, even with an exponential increase in the number of categories, pre-trained models remain pivotal in our fine-tuning strategy. The performance of fine-tuning consistently outperforms that of training from scratch.

However, regardless of whether the training approach is finetuning or training from scratch, a decline in performance is observed as the number of categories increases. The decline rates for SPD and DualPoseNet are slower, coupled with an initial augmentation in performance due to increased training data and iterations. In contrast, HS-Pose experiences a more rapid decline, with fine-tuned $5^\circ2~cm$ results dropping from initial 62.52\% to 14.42\%. Models that excel in tasks involving a limited number of categories may not necessarily maintain their superiority in large-vocabulary tasks, they might be surpassed by models that are more robust and easier to train.
\subsection{Visual Realism}
Due to the complexity of collecting and annotating real-world data, contemporary datasets like NOCS\cite{NOCS} are composed of a large amount of synthetic data and a small portion of real-world data. While collecting real data is relatively straightforward when the number of categories is limited, gathering well-annotated real-world data for pose estimation tasks involving large vocabulary categories becomes a monumental task.

\begin{figure}[t]
  \centering
   \includegraphics[width=0.90\linewidth]{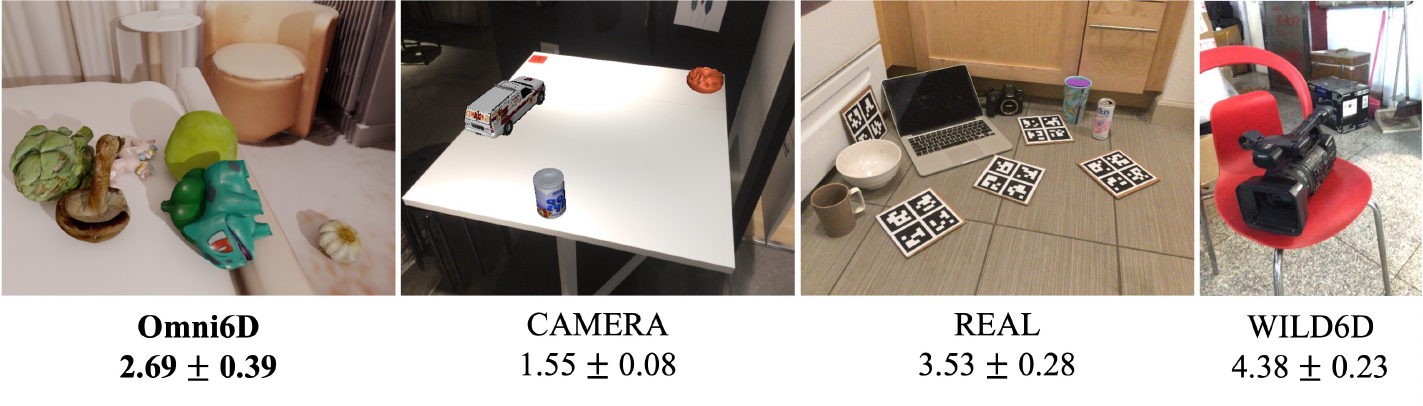}
   \caption{\textbf{Comparison of Visual Realism.} We evaluated the visual realism of Omni6D in comparison to other datasets through a survey involving 70 human subjects. We randomly selected 10 images from each dataset and introduced noise by blending in 5 images from COCO~\cite{coco}, which included captured photos, and SKETCH\protect\footnotemark, which comprised rendered images. Subjects were asked to rate the realism of sampled images on a scale from 1 (least realistic) to 5 (most realistic). We report the mean and standard deviation and include a sampled image from the study.}
   \label{fig:fidelity}
\end{figure}

Our Omni6D dataset, which includes large vocabulary objects, is also derived from rendering. However, the incorporation of real-scanned objects significantly enhances the realism of the rendered images. As depicted in~\cref{fig:fidelity}, Omni6D receives a score of $2.69 \pm 0.39$, surpassing the results obtained by CAMERA.

Given these significant advantages, our dataset excels not only in large-vocabulary scenarios but also in real-world scenes.
As depicted in~\cref{tab:real2sim}, We use DualPoseNet\cite{DualPoseNet} to train on the common categories in REAL~\cite{NOCS} and Omni6D, namely bottles, bowls, and mugs. We train separately on the two datasets and their mix. 
The results show that Omni6D models perform well on REAL275, and training on the mixed dataset outperforms using REAL or Omni6D datasets alone.
This demonstrates that our dataset enables the direct transfer of models to real-world scenes. Moreover, it seamlessly supplements the existing real-world dataset, enabling joint training of models on our dataset and the real-world data.

To further validate the sim2real capability of models trained with Omni6D, we constructed a real-world dataset, \textbf{\textit{Omni6D-Real}}, comprising 30 scenes, 39 categories, 73 instances, and 1k images. We captured RGBD images with Azure Kinect DK\footnote{https://learn.microsoft.com/azure/kinect-dk/} and preprocessed them using SAM~\cite{SAM} for object masks and ICP~\cite{ICP} for point cloud registration. Details are provided in Appendix Section D.
\footnotetext{https://sketchfab.com/}

\begin{table}[t]
    \caption{\textbf{Performance on REAL275 with Different Training Sets.} It compares how different training sets influence DualPoseNet's performance on REAL275~\cite{NOCS}, providing insights into the model's ability to generalize in real-world tasks using Omni6D.}
    \scriptsize
    \centering
    \begin{tabular}{@{}lc|cc|cccc|cc@{}}
    \toprule
        Train data  & Realism & $IoU_{50}$ & $IoU_{75}$ & $5^\circ2cm$ & $5^\circ5cm$ & $10^\circ2cm$ & $10^\circ5cm$ & $5^\circ$ & $2cm$ \\ 
          Omni6D & Real-Scanned & 78.76 & 32.69 & 6.55 & 8.80 & 15.00 & 21.38 & 11.20 & 49.54 \\ 
       REAL~\cite{NOCS} & Real& 84.51 & 43.43 & 8.76 & 10.40 & 21.24 & 25.39 & 13.01& 69.46\\ 
        REAL+Omni6D & Mixed  & \textbf{85.28} & \textbf{58.59} & \textbf{14.10} & \textbf{17.83} & \textbf{30.10} & \textbf{38.97} & \textbf{20.96} & \textbf{71.00}\\ 
        \bottomrule
    \end{tabular}
  \label{tab:real2sim}
\end{table}

\section{Conclusion}
\label{sec:discussion}
In conclusion, this paper introduces \textbf{\textit{Omni6D}}, a novel 6D pose estimation dataset with large-vocabulary categories and intricate scenes. We evaluate existing category-level 6D object pose estimation methods on this benchmark, analyze its challenges, and propose a fine-tuning strategy for large-vocabulary scenarios.

\noindent\textbf{Limitations.} 
Our dataset, though more complex, doesn't fully encompass all real-world challenges. Additionally, our fine-tuning strategy effectively extends methods from a small set to a larger one, but its efficacy may decrease with growing category diversity.

\noindent\textbf{Future Work.} 
Our study paves the way for diverse research avenues. An immediate next step is expanding the Omni6D dataset with more object types and scenes for comprehensive coverage. Additionally, annotating videos for scanned objects will validate algorithms' large-vocab pose estimation in real-world scenarios. Designing new training strategies for coping with increasing category diversity presents an intriguing challenge.



\title{Omni6D: Large-Vocabulary 3D Object Dataset for Category-Level 6D Object Pose Estimation\\
- Supplementary Materials -}
\titlerunning{Omni6D}

\author{}
\authorrunning{M.~Zhang et al.}

\institute{}

\maketitle

\renewcommand{\thesection}{\Alph{section}}
\setcounter{table}{0}
\setcounter{figure}{0}
\setcounter{footnote}{0}
\renewcommand{\thetable}{R\arabic{table}}
\renewcommand\thefigure{S\arabic{figure}}
\setcounter{page}{1}


\section{Overviews}
\label{sec:supplementary}
In the supplementary materials, we delve deeper into our research, offering a comprehensive exploration of several aspects mentioned in the main text. We unpack the details of the \textbf{\textit{Omni6D}} dataset, exploring its structure and statistics. We provide the construction details of the latest datasets, \textbf{\textit{Omni6D-xl}} and \textbf{\textit{Omni6D-Real}}. We provide a meticulous examination of the experimental procedures and analysis integral to our study. 
Additionally, we provided detailed insights into the questionnaire setting and result details regarding the visual realism of our Omni6D dataset.
These supplemental details are invaluable in facilitating a better understanding of our research methods and discoveries.

\begin{figure}[t]
  \centering
  \includegraphics[width=1\linewidth]{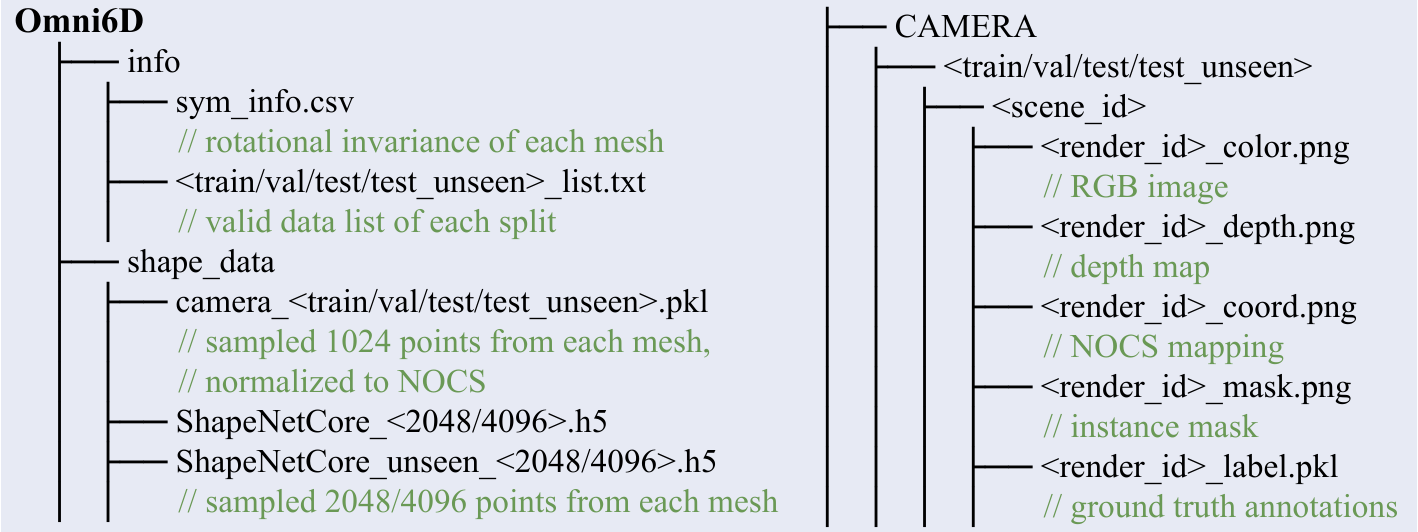}
   \caption{\textbf{Dataset structure.}}
   \label{fig:structure}
\end{figure}
\begin{table}[t]
  \centering
  \small
  \caption{\textbf{Detailed statistical overview of Omni6D dataset.} The table provides information about the number of categories, instances, and images in Omni6D$_{train}$, Omni6D$_{val}$, Omni6D$_{test}$ and Omni6D$_{out}$.}
  \begin{tabular}{@{}l|cc|cccc@{}}
    \toprule
        Datasets & \# Categories & \# Instances & \# Images \\ 
    \midrule
        Train & 166 & 3,294 & 812,602 \\ 
        Val & 166 & 919 & 28,661 \\ 
        Test & 166 & 475 & 14,267 \\ 
        Out & 17 & 52 & 4,762 \\ 
    \bottomrule
  \end{tabular}
  \label{tab:dataset-details}
\end{table}

\begin{figure}[t]
  \centering
   \includegraphics[width=0.5\linewidth]{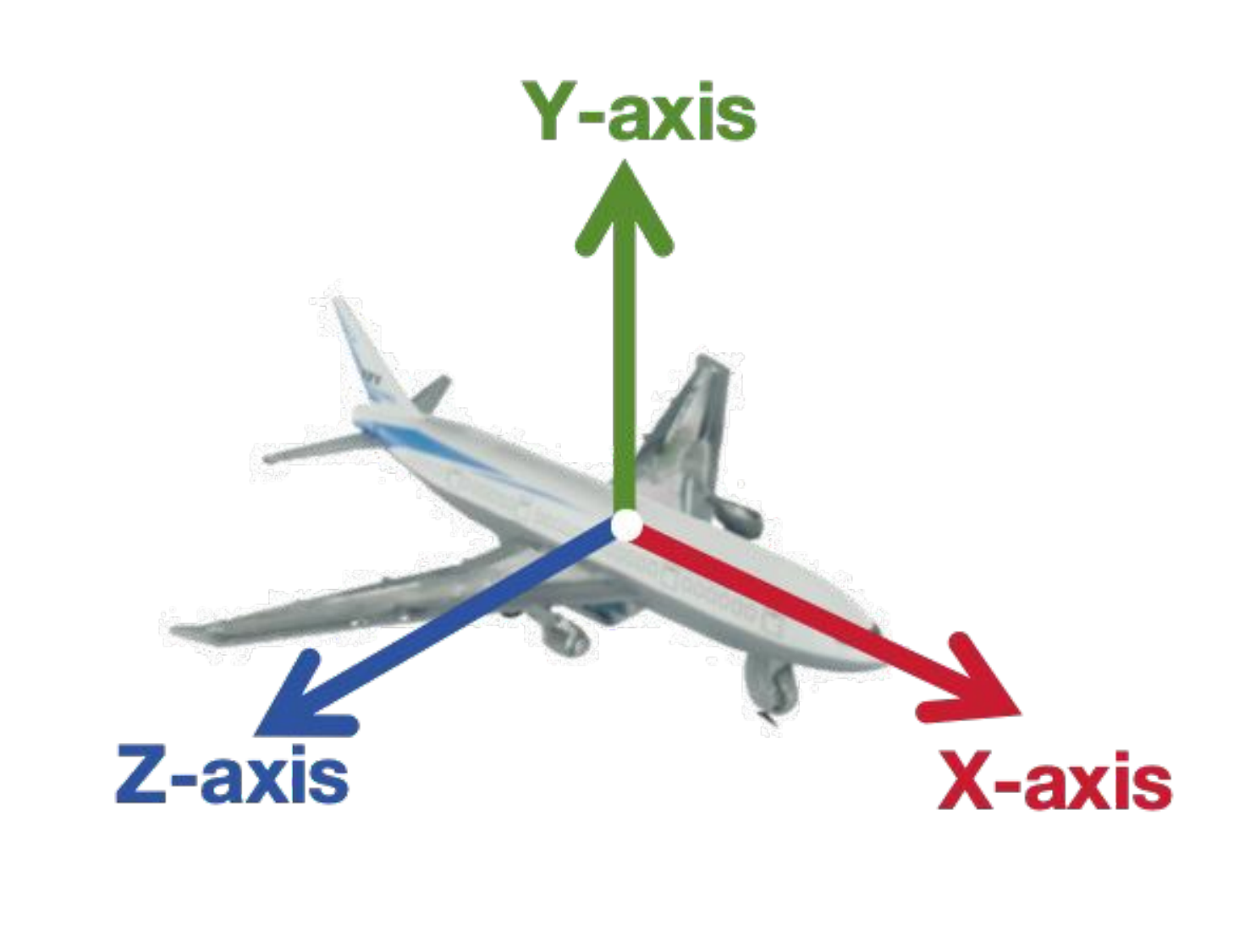}
   \caption{\textbf{An example instance adjusted to the canonical pose.} The canonical plane has its bottom-face normal aligned with -y and its front-face aimed at +x(akin to being upright and facing forward).}
   \label{fig:plane}
\end{figure}
\begin{figure}[t]
  \centering
   \includegraphics[width=0.8\linewidth]{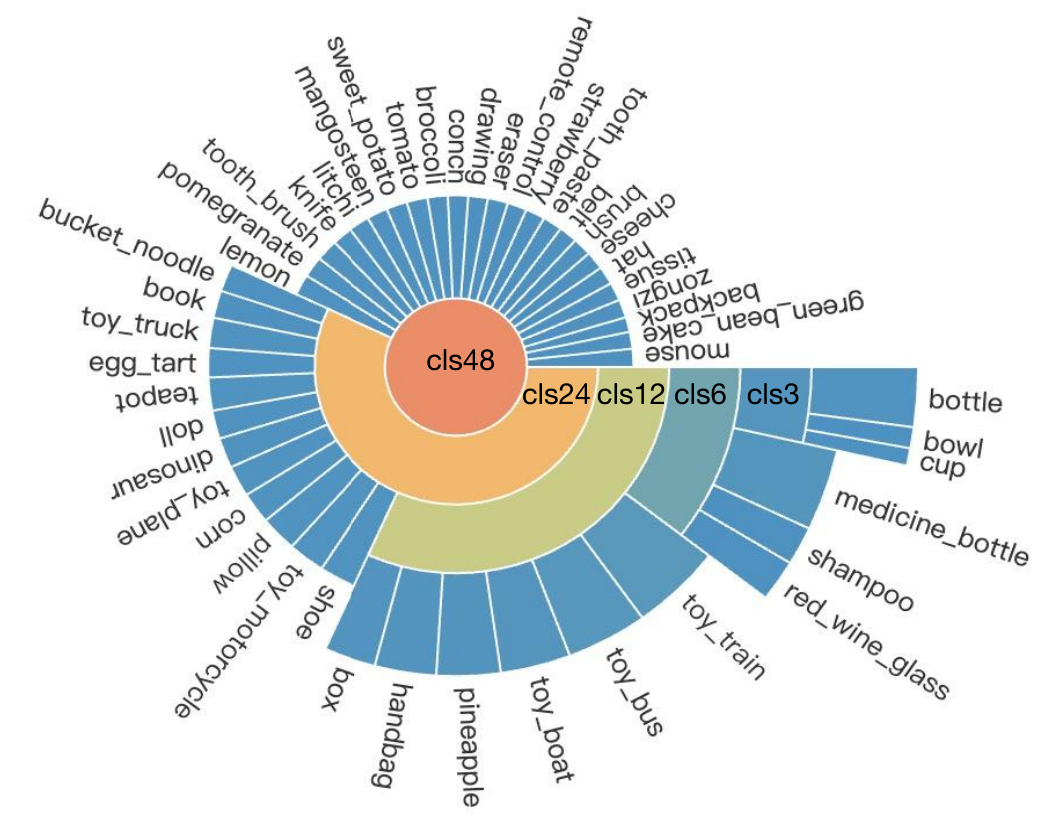}
   \caption{\textbf{Category inventory of cls$n$ within Omni6D.} The angle of each sector in the chart reflects the relative size of the instance count within that category.}
   \label{fig:cate_48}
\end{figure}
\begin{figure}[t]
  \centering
   \includegraphics[width=0.7\linewidth]{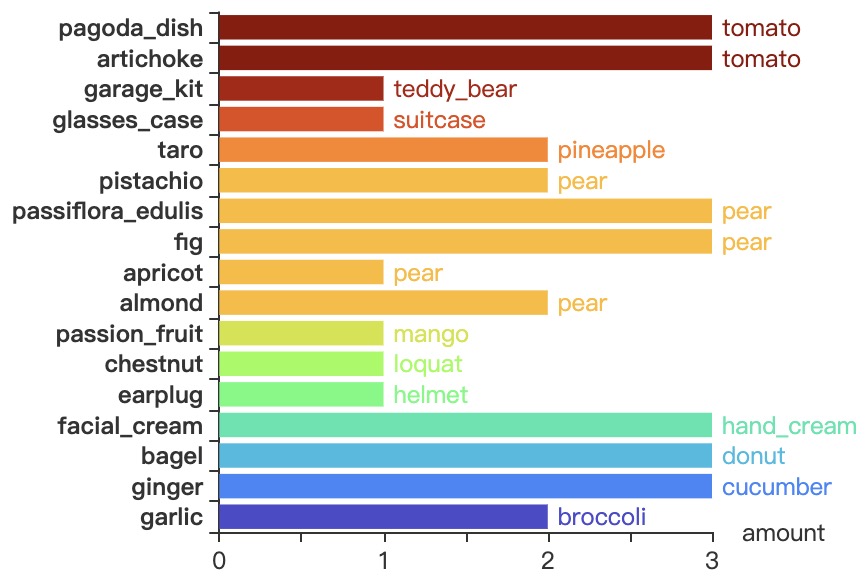}
   \caption{\textbf{Matching unseen categories from Omni6D$_{out}$ to Omni6D.} The unseen categories from Omni6D$_{out}$ are listed on the left side of the bar graph, while the matched known categories from Omni6D are displayed on the right, clearly illustrating the optimal correspondence between unseen and known categories based on cosine similarity. The horizontal axis displays the instance count for each corresponding category. Bars of the same color underscore the same match.}
   \label{fig:unseen}
\end{figure}

\section{Dataset Details}
\label{sec:Dataset-details}

\subsection{Omni6D overview}
\noindent\textbf{Dataset structure.} 
Our dataset is stored in folder-based structure. As illustrated in~\cref{fig:structure}, it comprises symmetry annotations, point clouds sampled from 3D scanned objects with adjusted canonical poses, and rendered views. We also provide a Blender-based simulation framework to facilitate users. 

Specifically for depth images, we applied a mapping transformation as mentioned in the main text. Original depth maps, saved as EXR files, have float32 precision with an accuracy of approximately $1e^{-7}$ and a size of 32 bits per pixel. Converting these depth maps to RGB format with a scaling factor of 10000 maintains a precision of about $1e^{-4}$, reducing storage size by 25\% with 24 bits per pixel. Due to PNG compression, actual storage can be reduced to 5\%-10\% of the original size. Also, our depth map compression method enables direct visualization in PNG format.

\noindent\textbf{Omni6D splits.} \cref{tab:dataset-details} provides information about the number of categories, instances, and images in Omni6D$_{train}$, Omni6D$_{val}$, Omni6D$_{test}$ and Omni6D$_{out}$. The categories are shared amongst the training, validation, and testing datasets, with a distribution ratio of 7:2:1 for instances. On the other hand, Omni6D$_{out}$ stands distinct, comprising an added set of 17 categories. Each split's images are exclusively derived from its corresponding instances, yet all splits share rendering parameters and backgrounds uniformly. To enable comprehensive model training, we have augmented the training set with an extensive volume of rendered images, reaching a total of 0.8M.

\noindent\textbf{Coordinate system.} We formulate a unified 3D coordinate system for all pose labels, positioning the camera center as the origin. In relation to the image captured, we set +x to face outward, +y to point upwards, and +z towards the left. The pose of an object is recorded relative to what we term a canonical pose object. As illustrated in \cref{fig:plane}, an instance adjusted to the canonical pose has its bottom-face normal aligned with -y and its front-face aimed at +x(akin to being upright and facing forward). The camera's intrinsic parameters are established as [577.5, 577.5, 319.5, 239.5], with the image size defined as 640 x 480 pixels. All data attributes, including details concerning the object's position and dimensions, are denoted in metric units.

\noindent\textbf{Diversity of scenes.} Each room is allocated a cube-shaped region, where objects are randomly positioned and fall free within room boundaries. Additionally, a lighting intensity range with a width of 2000 is established for each room model. 

\subsection{Omni6D Statistics}

We first provide a category inventory and corresponding instance counts for each category within Omni6D in \cref{fig:cate_166}. Most categories have [10, 50] objects.

In Section 4.1 of the main text, we mention cls$n$. Detailed categories from cls3 to cls48 are listed in \cref{fig:cate_48}. While subdividing the categories, we first select three categories that coincide with NOCS dataset~\cite{NOCS}, particularly those included in cls3: \textit{bottle}, \textit{bowl}, and \textit{cup}. Then, for cls6, we opt for three categories similar in shape to those in cls3, namely \textit{medicine\_bottle}, \textit{shampoo}, and \textit{red\_wine\_glass}. This selection aids in effectively finetuning the model across different categories. Following that, we generally select the remaining 42 categories based on the number of instances in each category, choosing from those with more instances to those with fewer.

\subsection{Omni6D$_{\text{out}}$ Statistics}

In Section 4.3 of the main text, we undertake 6D object pose estimation studies on Omni6D$_{out}$. This process begins by loading the pre-trained Word2Vec model \textit{GoogleNews-vectors-negative300.bin}. From the 166 categories available in Omni6D, we select the category that exhibits the highest cosine similarity with the unseen category for matching. As illustrated in \cref{fig:unseen}, the text to the right of the bar graph clarifies which categories are ultimately matched with the unseen category displayed on the left. For each unseen category, our model presumes its category as the one that is matched and proceeds with pose estimation accordingly. This visual representation provides an intuitive understanding of how our model leverages this matching information to predict the pose for each unseen category. Likewise, when evaluating the unseen categories, we also annotated the symmetrical information and implemented the metric processing as outlined in Section 4.2. 

\section{Omni6D-xl}
\label{sec:Omni6D-xl}

\begin{table}[t]
  \caption{\textbf{Comparisons between Omni6D, Omni6D-xl, Omni6D-Real and existing datasets.} Our datasets significantly extend the range of everyday object categories and instances.}
  \label{tab:omni6d_dataset}
  \centering
  \scriptsize
  \begin{tabular}{@{}l|cc|ccccc@{}}
    \toprule
    Datasets & Mode & Realism & \# Categories & \# Instances & \# Images \\
    \midrule
    ShapeNet-SRN Cars~\cite{inerf} &  RGB & Synthetic & 1 & 3514 & - \\ 
        Sim2Real Cars~\cite{inerf} &  RGB & Real & 1 & 10 & - \\
        \midrule
        CAMERA~\cite{NOCS} & RGBD & Synthetic & 6 & 1085 & 0.3M \\
        REAL~\cite{NOCS} & RGBD & Real & 6 & 42  & 8k \\
        Wild6D~\cite{reponet} & RGBD & Real & 5 & 1722 & 1M \\
        \midrule
        \textbf{Omni6D-Real} & RGBD & Real & 39 & 73 & 1k \\
        \textbf{Omni6D} & RGBD & Real-Scanned & 166 & 4,688 &0.8M \\
        \textbf{Omni6D-xl} & RGBD & Real-Scanned & \textbf{419} & \textbf{15,957} &1.1M \\
        
    \bottomrule
  \end{tabular}
\end{table}

Omni6D-xl extends Omni6D dataset by adding more categories and instance object models. Unlike normalizing all objects to the same scale, we retain the original scale of the objects and restore them to their actual size during rendering, adjusting other parameters accordingly. Moreover, we split our background rooms into training, validation, and test sets in a 2:1:1 ratio to avoid over-fitting on those scenes.

\noindent\textbf{Dataset Collection.} As shown in~\cref{tab:omni6d_dataset}, Omni6D-xl comprises 15,957 instances across an impressive span of 419 categories, with 15,474 instances across 319 categories used as the train/valid/test dataset. Additionally, 483 instances across 100 unseen categories are used to assess the model's inter-category generalization capabilities.
Each instance is a high-resolution textured mesh, obtained using Shining 3D scanner\footnote{https://www.einscan.com/} and Artec Eva 3D scanner\footnote{https://www.artec3d.cn/}, collected from OmniObject3D~\cite{oo3d}. We normalize object models to fit within a $(-1,1)^3(m^3)$ three-dimensional space, and align objects within each category to a consistent canonical pose. Additionally, we store the scale of the object models.

\noindent\textbf{Rendering.} We employ stratified sampling to split instances within each category, subsequently dividing them into training, validation, and test sets in a 8:1:1 ratio. 
In constructing our dataset, we utilize 8 room models from the Replica dataset as backdrops, splitting them into training, validation, and test sets in a 2:1:1 ratio.
For each scenery setup, we randomly select a room model to act as the background, along with 4-6 object instance models.
Each room is allocated a cube-shaped region where objects are randomly positioned and allowed to fall freely within room boundaries, resulting in random scattering in a specific section of the room.
Additionally, a lighting intensity range with a width of 2000 is established for each room model.
Each object model is scaled by the pre-stored scale factor divided by 50. Considering the attention center of the combined instance models as the origin point, the camera randomly selects ten positions within an elevation angle range between $30-90^\circ$. The camera then performs rendering at these selected positions while facing towards the attention center.

\noindent\textbf{Setting.} We utilize BlenderProc 2.5.0~\cite{blenderproc} to implement the aforementioned rendering process. The intrinsic parameters of the camera are set to [577.5, 577.5, 319.5, 239.5], with an image size specified as $640\times480$. Our approach ensures the diversity and breadth of the dataset, making it suitable for rigorous testing and yielding accurate results.

\section{Omni6D-Real}
\label{sec:Omni6D-Real}
To further validate the sim2real capability of models trained with Omni6D and reduce the gap between our dataset and real-world data, we constructed a real-world dataset, \textbf{\textit{Omni6D-Real}}. As shown in~\cref{tab:omni6d_dataset}, it comprises 30 scenes, 39 categories, 73 instances, and 1k images.

\begin{figure}[t]
  \centering
  \includegraphics[width=\linewidth]{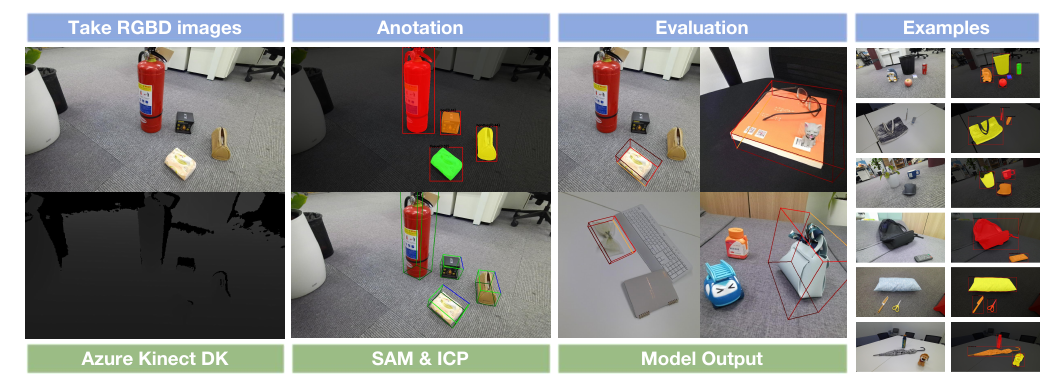}
   \caption{\textbf{Constructing Omni6D-Real: pipeline \& examples.}}
   \label{fig:pipeline}
\end{figure}

\noindent\textbf{Dataset Construction.} As shown in~\cref{fig:pipeline}, we captured RGBD images with the Azure Kinect DK\footnote{https://learn.microsoft.com/azure/kinect-dk/} and preprocessed them using SAM~\cite{SAM} for object masks and ICP~\cite{ICP} for point cloud registration. The intrinsic parameters of the camera are set to [605.81, 605.63, 641.72, 363.23], with an image size specified as $1280\times720$.
For each scene, we manually annotated 3D bounding boxes for the first frame and derived bboxes for the next frame based on registered poses. Addressing the inherent limitations of ICP, particularly its accumulating errors, we further refined the derived bboxes through manual adjustments. This iterative process, where ICP serves as an aid to manual annotation, ensures the accuracy of 3D bboxes across all frames.

\noindent\textbf{Evaluation.} We evaluated the performance of DualPoseNet~\cite{DualPoseNet} on our processed real-world dataset. Despite being trained solely on simulated data, the model exhibited excellent performance on real-world tasks. This demonstrates to a certain extent that our real-scanned 3D models can minimize the gap between synthetic and real images.

\section{Additional Experimental Details}
\label{sec: Experiements Details}

\subsection{Experimental Settings}
\begin{table}[t]
  \centering
  \small
  \caption{\textbf{Detailed parameters.} Experimental settings on different baselines.}
  \begin{tabular}{@{}l|ccc@{}}
    \toprule
        Model& Learning\_rate  & Batch\_size & \# GPUs \\ 
    \midrule
        SPD~\cite{SPD} & 1e-4 & 128 & 4\\ 
        SGPA~\cite{SGPA} & 1e-4 & 128 & 4\\ 
        DualPoseNet~\cite{DualPoseNet} & 1e-4 & 128 & 1\\ 
        RBP-Pose~\cite{RBP} & 1e-4 & 256 & 4\\ 
        GPV-Pose~\cite{GPV} & 1e-4 & 256 & 4\\ 
        HS-Pose~\cite{HS} & 1e-4 & 256 & 4\\ 
    \bottomrule
  \end{tabular}
  \label{tab:para}
\end{table}
\begin{table}[t]
    \small
    \centering
    \caption{\textbf{Performance of top-20 categories on Omni6D.} Models are trained on Omni6D$_{train}$ and tested on Omni6D$_{test}$. The table demonstrates the average performance of each algorithm across the top 20 categories, as measured by the $5^\circ2cm$ metric. \textbf{Bold} and \underline{underlined} results indicate the best and second-best performers.}
    \resizebox{\textwidth}{!}{%
    \begin{tabular}{@{}lc|cc|cccc|cccc@{}}
    \toprule
        Methods & Network & $IoU_{50}$ & $IoU_{75}$ & $5^\circ2cm$ & $5^\circ5cm$ & $10^\circ2cm$ & $10^\circ5cm$ & $5^\circ$ & $10^\circ$ & $2cm$ & $5cm$ \\ 
        \midrule
        SPD~\cite{SPD} & implicit & 68.65 & \underline{45.27} & \textbf{24.19} & \textbf{26.78} & \underline{37.18} & \textbf{42.15} & \textbf{27.60} & \textbf{43.34} & 60.49 & 84.03 \\
        SGPA~\cite{SGPA} & implicit & 70.40 & \textbf{48.23} & \underline{20.17} & \underline{21.79} & \textbf{37.30} & \underline{40.78} &\underline{ 22.26} & \underline{41.87} & 63.39 & 86.14 \\
        DualPoseNet~\cite{DualPoseNet} & hybrid & \textbf{74.09} & 41.50 & 15.56 & 17.11 & 30.25 & 32.78 & 17.14 & 32.84 & \underline{83.48} & \textbf{98.46} \\
        RBP-Pose~\cite{RBP} & hybrid & 42.03 & 10.74 & 2.84 & 4.54 & 3.41 & 5.21 & 4.68 & 5.37 & 44.77 & 88.43 \\
        GPV-Pose~\cite{GPV} & explicit & 19.53 & 0.78 & 0.74 & 3.17 & 0.81 & 3.58 & 7.06 & 7.85 & 7.03 & 39.86 \\
        HS-Pose~\cite{HS} & explicit & \underline{72.36} & 37.81 & 11.94 & 13.26 & 22.08 & 23.93 & 13.37 & 24.07 & \textbf{86.11} & \underline{98.31} \\ 
        \bottomrule
    \end{tabular}
    }
  \label{tab:best_20}
\end{table}
\begin{figure*}[t]
  \centering
   \includegraphics[width=0.94\linewidth]{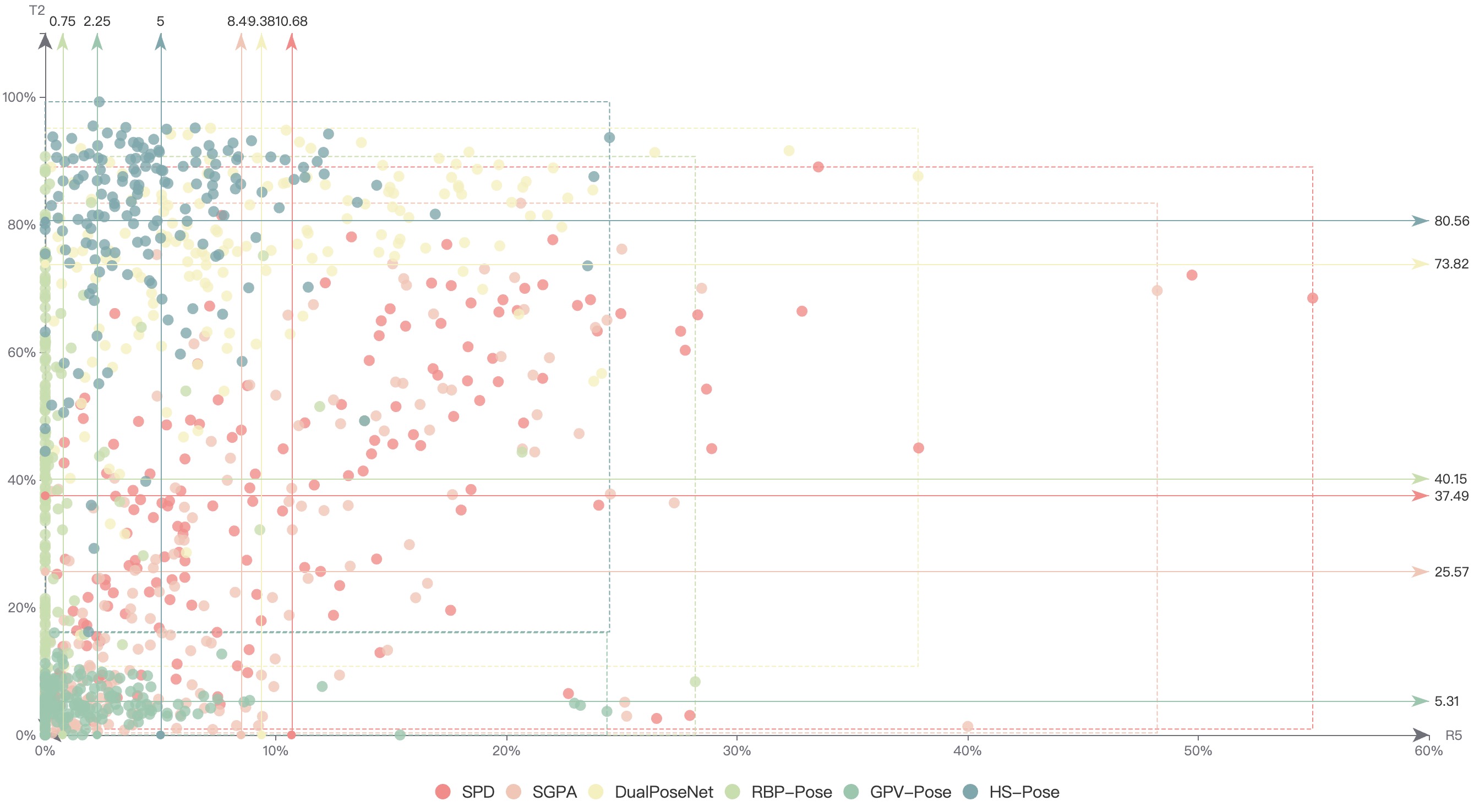}
   \caption{\textbf{Metrics $5^{\circ}$ and $2~cm$ results on Omni6D categories.} It showcases the $5^{\circ}$~(R5) and $2~cm$~(T2) metrics for various models across different categories on the Omni6D test set. Each color represents a model, with each point indicating a category result. Dashed lines outline the range of each model's $5^{\circ}$~(R5) and $2~cm$~(T2) metrics, while arrows depict their means.}
   \label{fig:supply_cate}
\end{figure*}
\begin{table}[t]
    \small
    \centering
    \caption{\textbf{Non-symmetry-aware metric results on Omni6D.} Models are trained on Omni6D$_{train}$ and tested on Omni6D$_{test}$, while not using our symmetry-aware metric.}
    \begin{tabular}{@{}l|cc|cccc|cccc@{}}
    \toprule
        Methods &$IoU_{50}$ & $IoU_{75}$ & $5^\circ2cm$ & $5^\circ5cm$ & $10^\circ2cm$ & $10^\circ5cm$ & $5^\circ$ & $10^\circ$ & $2cm$ & $5cm$ \\ 
        \midrule
        SPD~\cite{SPD} & 30.82 &\textbf{ 13.09} & \textbf{3.36} & \textbf{3.62} & \textbf{8.10} & \textbf{9.06} & \textbf{3.65} & \textbf{9.19} & 38.32 & 71.43\\ 
        SGPA~\cite{SGPA} & 26.43 & 10.06 & \underline{2.34} & \underline{2.57 }& 6.25 & \underline{7.40} & \underline{2.59} & \underline{7.62} & 26.11 & 60.67\\ 
        DualPoseNet~\cite{DualPoseNet} & \underline{35.78} & \underline{12.32} & 2.06 & 2.11 & \underline{6.47} & 6.74 & 2.11 & 6.75 & \underline{74.13} & \underline{96.42}\\
        RBP-Pose~\cite{RBP} & 14.77 & 0.63 & 0.00 & 0.00 & 0.00 & 0.01 & 0.00 & 0.01 & 34.33 & 73.54 \\
        GPV-Pose~\cite{GPV} & 5.50 & 0.02 & 0.00 & 0.01 & 0.01 & 0.04 & 0.02 & 0.07 & 5.37 & 33.31\\
        HS-Pose~\cite{HS} & \textbf{39.18} & 9.68 & 0.36 & 0.37 & 2.30 & 2.43 & 0.37 & 2.44 & \textbf{80.65} & \textbf{97.64} \\ 
        \bottomrule
    \end{tabular}
  \label{tab:nosym166}
\end{table}
\begin{table*}[!h]
    \small
    \centering
    \caption{\textbf{Individual category performance on unseen categories. } Models are trained on Omni6D$_{train}$ and tested on Omni6D$_{out}$, using the optimal DualPoseNet~\cite{DualPoseNet} model. The table distinctly presents results for each category, with the 1st column representing the category name and the 2nd column indicating the corresponding known matched category. The table is sorted in descending order based on the metric $5^\circ2cm$.}
    \resizebox{\textwidth}{!}{%
    \begin{tabular}{@{}lc|cc|cccc|cccc@{}}
    \toprule
        Category & Match& $IoU_{50}$ & $IoU_{75}$ & $5^\circ2cm$ & $5^\circ5cm$ & $10^\circ2cm$ & $10^\circ5cm$ & $5^\circ$ & $10^\circ$ & $2cm$ & $5cm$ \\ 
        \midrule
        passion\_fruit & mango & 58.79 & 25.08 & 8.44 & 8.77 & 18.51 & 18.99 & 8.77 & 19.48 & 85.23 & 99.19 \\ 
        facial\_cream & hand\_cream & 53.43 & 28.61 & 7.58 & 7.82 & 16.38 & 17.60 & 7.82 & 17.60 & 84.11 & 96.82 \\ 
        taro & pineapple & 58.70 & 26.48 & 5.50 & 5.65 & 16.49 & 16.79 & 5.65 & 16.79 & 89.47 & 99.39 \\ 
        fig & pear & 60.64 & 24.79 & 3.52 & 3.63 & 15.69 & 15.90 & 3.95 & 16.33 & 84.85 & 99.68 \\ 
        garlic & broccoli & 32.85 & 4.78 & 3.18 & 3.18 & 6.16 & 6.58 & 3.18 & 6.79 & 81.95 & 99.79 \\ 
        earplug & helmet & 35.04 & 14.14 & 2.69 & 3.08 & 9.23 & 9.87 & 3.08 & 9.87 & 88.72 & 98.72 \\ 
        passiflora\_edulis & pear & 37.76 & 15.22 & 1.82 & 1.82 & 7.29 & 7.75 & 1.82 & 7.75 & 77.51 & 99.24 \\ 
        bagel & donut & 38.71 & 13.92 & 1.75 & 2.25 & 9.64 & 10.26 & 2.25 & 10.26 & 83.35 & 99.12 \\ 
        artichoke & tomato & 48.20 & 13.77 & 0.78 & 0.90 & 3.70 & 4.26 & 1.01 & 4.48 & 79.28 & 99.22 \\ 
        pagoda\_dish & tomato & 32.66 & 4.45 & 0.78 & 0.91 & 2.22 & 2.61 & 0.91 & 3.00 & 82.40 & 98.31 \\ 
        ginger & cucumber & 30.33 & 1.54 & 0.78 & 0.78 & 2.08 & 2.86 & 0.78 & 2.86 & 59.48 & 98.70 \\ 
        almond & pear & 23.43 & 1.88 & 0.76 & 0.89 & 1.78 & 2.28 & 1.02 & 2.41 & 80.58 & 99.11 \\ 
        garage\_kit & teddy\_bear & 26.66 & 5.44 & 0.59 & 0.73 & 2.42 & 2.87 & 0.83 & 3.08 & 71.59 & 97.09 \\ 
        glasses\_case & suitcase & 37.99 & 5.24 & 0.44 & 0.44 & 0.88 & 1.32 & 0.44 & 1.32 & 90.79 & 96.93 \\ 
        chestnut & loquat & 22.73 & 1.97 & 0.27 & 0.40 & 0.93 & 1.19 & 0.40 & 1.19 & 57.43 & 97.75 \\ 
        pistachio & pear & 23.56 & 2.03 & 0.17 & 0.17 & 0.34 & 0.52 & 0.17 & 0.69 & 80.72 & 99.66 \\ 
        apricot & pear & 11.48 & 0.26 & 0.00 & 0.00 & 0.00 & 0.00 & 0.00 & 0.00 & 80.32 & 100.00 \\ 
        \bottomrule
    \end{tabular}
    }
  \label{tab:unseen}
\end{table*}

All experiments are conducted on a server equipped with 96 Intel(R) Xeon(R) Gold 6248R CPUs @ 3.00GHz and 8 NVIDIA A100-SXM4-80GB GPUs. We ensure consistency in all parameters and strategies throughout training, thereby maintaining uniformity in our experimental environment. For our baseline model, we adhere to the same parameters as provided by the original authors, with modifications only made to \textit{learning\_rate}, \textit{batch\_size}, and the corresponding number of GPUs used. Detailed parameters are displayed in \cref{tab:para}.

We encountered some challenges during model training. 
Due to the larger batch size we selected compared to the original model, the training speed of the GPV-Pose model became excessively slow. The main reason for this issue is that GPV-Pose~\cite{GPV} model uses ``for loop'' for batch processing during training, which is inefficient when dealing with large-scale data. We optimized the model by replacing ``for loop'' with batch computations carried out at the Tensor level. This modification significantly accelerated our training speed, effectively ensuring the efficient functioning of the model.


\subsection{Performance on Omni6D}
In this section, we provide the results of the $5^\circ$ and $2~cm$ metrics for categories in Omni6D. \cref{fig:supply_cate} showcases the $5^{\circ}$(R5) and $2~cm$(T2) metrics for various models across different categories on the Omni6D test set. The results show that SPD and SGPA excel particularly in predicting rotations, potentially due to their implicit networks' tendency to generate more accurate rotational predictions. On the other hand, DualPoseNet, HS-Pose and RBP-Pose offer superior estimates for translations, likely related to the capabilities of explicit network models to deliver better translation and size estimations. These findings further affirm the speculations made in Section 4.3.

\cref{tab:best_20} demonstrates the average performance of each algorithm across the top 20 categories, as measured by the $5^\circ2~cm$ metric. As shown in the table, it's evident that all algorithms show improved performance across various metrics compared to the full set of 166 categories, which is foreseeable. While all algorithms see similar improvements, SPD and SGPA stand out with notable progress. Considering their bad performance on unseen categories, as outlined in the main text, it's clear that they exhibit considerable variability in predictive accuracy across different categories. This suggests that SPD and SGPA employ a nuanced approach, finetuning their strategies for each category by leveraging their implicit network methodologies. These methodologies sync well with specific features and challenges of certain categories, enabling more accurate predictions. Conversely, their effectiveness lessens when applied to categories that mismatch their methodologies.

We also report the non-symmetry-aware metric results in \cref{tab:nosym166}, showing a notable performance drop compared to the symmetry-aware metric presented in Tab.~2. As discussed in Fig.~2, the prevalence of rotational invariance in 3D models makes the consideration of symmetry indispensable.

\subsection{Generalization Performance}
\cref{tab:unseen} distinctly presents the results for each category, derived from tests using the optimal DualPoseNet~\cite{DualPoseNet} model. In this table, the first column lists the category name while the second column indicates the corresponding known matched category.
It can be observed that prediction for translation is almost category-independent, while rotation is closely related to the category.
\subsection{Category-wise Analysis}
In the corresponding subsection under Section 4.3, we introduce the concept of diversity. Assume that $C_i$ is the set of all instances within category $i$, $c_{ij}$ and $c_{ik}$ are two instances within this set, and $\text{{Chamfer}}(c_{ij}, c_{ik})$ is the Chamfer distance~\cite{chamfer} between instances $c_{ij}$ and $c_{ik}$. Then, the diversity $D_i$ within category $i$ can be calculated as:

\begin{equation}
D_i = \frac{1}{|C_i|^2} \sum_{j=1}^{|C_i|} \sum_{k=1}^{|C_i|} \text{{Chamfer}}(c_{ij}, c_{ik}).
\end{equation}

Essentially, this formula calculates the average Chamfer distance among all possible pairs of instances within a category, serving as a measure of diversity for that category. A larger result indicates higher intra-class diversity among instances within that category. \cref{fig:diversity} depicts the intra-class diversity across various categories in Omni6D.

\begin{table*}[t]
    \small
    \centering
    \caption{\textbf{Performance of SPD on Omni6D dataset trained from scratch.} It presents the performance of the SPD model when trained from scratch separately on various subsets of the Omni6D dataset, specifically cls3, cls6, cls12, cls24, and cls48, each of which contains a different number of categories.}
    \resizebox{\textwidth}{!}{%
    \begin{tabular}{@{}ll|cc|cccc|cccc@{}}
    \toprule
        Train & Test & $IoU_{50}$ & $IoU_{75}$ & $5^\circ2cm$ & $5^\circ5cm$ & $10^\circ2cm$ & $10^\circ5cm$ & $5^\circ$ & $10^\circ$ & $2cm$ & $5cm$ \\ 
        \midrule
        train-from-scratch(cls3) & cls3 & 44.27 & 20.50 & 9.52 & 10.56 & 14.45 & 17.13 & 10.85 & 17.98 & 41.98 & 65.42 \\ 
        train-from-scratch(cls6) & cls6 & 54.94 & 28.37 & 14.96 & 16.86 & 20.75 & 24.91 & 17.26 & 25.62 & 51.13 & 74.89 \\ 
        train-from-scratch(cls12) & cls12 & 55.30 & 29.47 & 12.92 & 15.01 & 21.90 & 26.31 & 15.59 & 27.58 & 49.99 & 77.51 \\ 
        train-from-scratch(cls24) & cls24 & 57.37 & 31.08 & 11.90 & 13.44 & 22.67 & 26.36 & 14.02 & 27.62 & 52.98 & 79.73 \\ 
        train-from-scratch(cls48) & cls48 & 48.22 & 24.54 & 9.07 & 11.07 & 17.53 & 22.15 & 11.89 & 23.63 & 41.60 & 73.28 \\ 
        \bottomrule
    \end{tabular}
    }
  \label{tab:direct-SPD}
\end{table*}
\begin{table*}[t]
    \small
    \centering
    \caption{\textbf{Performance of SPD on Omni6D dataset with finetuning strategy.} It presents the performance of the SPD model initially pretrained on CAMERA dataset~\cite{NOCS} and then incrementally finetuned using various subsets of the Omni6D dataset, specifically cls3, cls6, cls12, cls24, and cls48.}
    \resizebox{\textwidth}{!}{%
    \begin{tabular}{@{}ll|cc|cccc|cccc@{}}
    \toprule
        Train & Test &$IoU_{50}$ & $IoU_{75}$ & $5^\circ2cm$ & $5^\circ5cm$ & $10^\circ2cm$ & $10^\circ5cm$ & $5^\circ$ & $10^\circ$ & $2cm$ & $5cm$ \\ 
        \midrule
        pretrain~(CAMERA) & cls3 & 16.63 & 0.79 & 0.05 & 0.59 & 0.09 & 0.93 & 2.55 & 3.60 & 2.29 & 23.53 \\ 
        finetune~(CAMERA+cls3) & cls3 & 46.19 & 21.42 & 10.20& 11.49 & 16.71 & 19.59 & 11.79 & 20.18 & 53.39 & 79.83 \\ 
        finetune~(CAMERA+cls6) & cls6 & 60.85 & 32.57 & 15.09 & 17.84 & 23.63 & 28.54 & 18.02 & 28.82 & 62.89 & 86.03 \\ 
        finetune~(CAMERA+cls12) & cls12 & 56.67 & 29.71 & 13.18 & 15.08 & 22.54 & 26.34 & 15.50 & 26.92 & 58.31 & 83.68 \\ 
        finetune~(CAMERA+cls24) & cls24 & 55.82 & 28.81 & 12.06 & 13.58 & 22.24 & 26.02 & 14.03 & 26.95 & 57.92 & 84.36 \\ 
        finetune~(CAMERA+cls48) & cls48 & 45.06 & 22.76 & 9.22 & 11.22 & 17.10 & 21.36 & 11.96 & 22.67 & 44.80 & 74.66 \\ 
        \bottomrule
    \end{tabular}
    }
  \label{tab:finetune-SPD}
\end{table*}
\begin{table*}[t]
    \small
    \centering
    \caption{\textbf{Performance of DualPoseNet on Omni6D trained from scratch.} It presents the performance of the DualPoseNet model when trained from scratch separately on various subsets of Omni6D.}
    \resizebox{\textwidth}{!}{%
    \begin{tabular}{@{}ll|cc|cccc|cccc@{}}
    \toprule
        Train & Test & $IoU_{50}$ & $IoU_{75}$ & $5^\circ2cm$ & $5^\circ5cm$ & $10^\circ2cm$ & $10^\circ5cm$ & $5^\circ$ & $10^\circ$ & $2cm$ & $5cm$ \\ 
        \midrule
        train-from-scratch(cls3) & cls3 & 66.53 & 39.60 & 17.03 & 17.24 & 29.20 & 29.68 & 17.24 & 29.68 & 90.29 & 96.60 \\ 
        train-from-scratch(cls6) & cls6 & 76.27 & 44.62 & 20.59 & 21.12 & 32.59 & 33.84 & 21.16 & 33.90 & 87.75 & 96.81 \\ 
        train-from-scratch(cls12) & cls12 & 68.21 & 37.83 & 17.06 & 18.02 & 27.86 & 29.70 & 18.08 & 29.79 & 81.73 & 96.52 \\ 
        train-from-scratch(cls24) & cls24 & 70.14 & 43.01 & 19.99 & 20.92 & 33.03 & 34.83 & 21.03 & 34.94 & 82.58 & 96.90 \\ 
        train-from-scratch(cls48) & cls48 & 65.00 & 33.47 & 10.18 & 11.07 & 23.43 & 25.50 & 11.12 & 25.63 & 76.38 & 96.48 \\ 
        \bottomrule
    \end{tabular}
    }
  \label{tab:direct-Dual}
\end{table*}
\begin{table*}[!h]
    \small
    \centering
    \caption{\textbf{Performance of DualPoseNet on Omni6D with finetuning strategy.} It presents the performance of the DualPoseNet model initially pretrained on CAMERA dataset~\cite{NOCS} and then incrementally finetuned using various subsets of Omni6D.}
    \resizebox{\textwidth}{!}{%
    \begin{tabular}{@{}ll|cc|cccc|cccc@{}}
    \toprule
        Train & Test &$IoU_{50}$ & $IoU_{75}$ & $5^\circ2cm$ & $5^\circ5cm$ & $10^\circ2cm$ & $10^\circ5cm$ & $5^\circ$ & $10^\circ$ & $2cm$ & $5cm$ \\ 
        \midrule
        pretrain~(CAMERA) & cls3 & 29.04 & 5.45 & 3.42 & 3.91 & 3.92 & 4.48 & 4.05 & 4.62 & 67.28 & 89.58 \\ 
        finetune~(CAMERA+cls3) & cls3 & 75.25 & 44.57 & 17.72 & 17.72 & 32.95 & 33.70 & 17.72 & 33.70 & 91.52 & 96.51 \\ 
        finetune~(CAMERA+cls6) & cls6 & 77.34 & 46.32 & 23.17 & 23.66 & 34.00 & 35.27 & 23.73 & 35.37 & 89.96 & 97.32 \\ 
        finetune~(CAMERA+cls12) & cls12 & 68.61 & 37.58 & 17.17 & 17.88 & 28.15 & 29.45 & 17.94 & 29.58 & 83.22 & 96.83 \\ 
        finetune~(CAMERA+cls24) & cls24 & 70.68 & 43.00 & 22.55 & 22.96 & 33.82 & 35.61 & 22.96 & 35.61 & 89.62 & 96.83 \\ 
        finetune~(CAMERA+cls48) & cls48 & 64.60 & 34.52 & 13.57 & 14.36 & 25.34 & 27.01 & 14.45 & 27.20 & 77.68 & 96.08 \\ 
        \bottomrule
    \end{tabular}
    }
  \label{tab:finetune-Dual}
\end{table*}
\begin{table*}[t]
    \small
    \centering
    \caption{\textbf{Performance of HS-Pose on Omni6D trained from scratch.} It presents the performance of the HS-Pose model when trained from scratch separately on various subsets of Omni6D.}
    \resizebox{\textwidth}{!}{%
    \begin{tabular}{@{}ll|cc|cccc|cccc@{}}
    \toprule
        Train & Test & $IoU_{50}$ & $IoU_{75}$ & $5^\circ2cm$ & $5^\circ5cm$ & $10^\circ2cm$ & $10^\circ5cm$ & $5^\circ$ & $10^\circ$ & $2cm$ & $5cm$ \\ 
        \midrule
        train-from-scratch(cls3) & cls3 & 94.46 & 86.57 & 47.65 & 48.28 & 81.81 & 83.70 & 48.54 & 83.96 & 90.33 & 97.21 \\ 
        train-from-scratch(cls6) & cls6 & 93.61 & 82.65 & 48.79 & 49.93 & 74.03 & 76.24 & 49.93 & 76.33 & 90.71 & 97.68 \\ 
        train-from-scratch(cls12) & cls12 & 81.40 & 57.79 & 21.78 & 22.13 & 42.79 & 43.93 & 22.13 & 43.94 & 87.48 & 98.45 \\ 
        train-from-scratch(cls24) & cls24 & 79.75 & 52.25 & 16.92 & 17.58 & 37.17 & 38.93 & 17.59 & 38.95 & 87.66 & 98.38 \\ 
        train-from-scratch(cls48) & cls48 & 73.30& 39.62 & 8.79 & 9.16 & 23.97 & 25.41 & 9.18 & 25.49 & 83.69 & 98.42 \\ 
        \bottomrule
    \end{tabular}
    }
  \label{tab:direct-hspose}
\end{table*}
\begin{table}[t]
    \small
    \centering
    \caption{\textbf{Performance of HS-Pose on Omni6D with finetuning strategy.} It presents the performance of the HS-Pose model initially pretrained on CAMERA dataset~\cite{NOCS} and then incrementally finetuned using various subsets of Omni6D.}
    \resizebox{\textwidth}{!}{%
    \begin{tabular}{@{}ll|cc|cccc|cccc@{}}
    \toprule
        Train & Test &$IoU_{50}$ & $IoU_{75}$ & $5^\circ2cm$ & $5^\circ5cm$ & $10^\circ2cm$ & $10^\circ5cm$ & $5^\circ$ & $10^\circ$ & $2cm$ & $5cm$ \\ 
        \midrule
        pretrain~(CAMERA) & cls3 & 31.67 & 7.14 & 4.13 & 5.12 & 6.38 & 7.91 & 5.19 & 8.25 & 70.56 & 92.27 \\ 
        finetune~(CAMERA+cls3) & cls3 & 94.04 & 88.29 & 62.52 & 63.92 & 84.51 & 87.41 & 63.92 & 87.41 & 90.87 & 97.60 \\ 
        finetune~(CAMERA+cls6) & cls6 & 94.47 & 86.19 & 56.20& 58.06 & 79.82 & 82.76 & 58.10& 82.89 & 90.76 & 97.87 \\ 
        finetune~(CAMERA+cls12) & cls12 & 83.85 & 61.37 & 28.37 & 29.03 & 48.73 & 50.37 & 29.03 & 50.43 & 87.48 & 97.98 \\ 
        finetune~(CAMERA+cls24) & cls24 & 81.35 & 56.59 & 23.27 & 24.04 & 43.22 & 45.16 & 24.05 & 45.22 & 87.68 & 98.43 \\ 
        finetune~(CAMERA+cls48) & cls48 & 75.18 & 45.11 & 14.42 & 15.02 & 30.45 & 32.24 & 15.02 & 32.31 & 83.34 & 98.45 \\  
        \bottomrule
    \end{tabular}
    }
  \label{tab:finetune-hspose}
\end{table}

\subsection{Finetune from Limited Categories}
As elaborated in the corresponding subsection under Section 4.3 in the main text, \cref{tab:direct-SPD,tab:finetune-SPD,tab:direct-Dual,tab:finetune-Dual,tab:direct-hspose,tab:finetune-hspose} respectively present the specific numerical results of the training from scratch and finetuning experiments conducted by SPD, DualPoseNet, and HS-Pose.

For the training from scratch experiments, it is observed that an increase in the number of categories during the training and testing phases generally leads to a decline in most performance indicators. Contrastingly, in the finetuning experiments, as the number of categories used for finetuning and testing increases, most performance indicators do show a decline. However, certain metrics like $5~cm$ remain relatively stable, and the decrease in other metrics isn't as severe as when training from scratch. This observation points to the robustness of the pretraining and incremental finetuning approach across a different number of categories, emphasizing its effectiveness.


\subsection{Qualitative Comparisons}
For category-level 6D pose and size estimation, we visualize more qualitative results of different methods on Omni6D$_{test}$ and Omni6D$_{out}$ in \cref{fig:qualitative} and \cref{fig:qualitative_unseen}. These figures illustrate the models' ability to generalize within known categories (intra-class generalization) as well as across unseen categories (inter-class generalization).

\section{Visual Realism}
\subsection{Questionnaire settings}
We evaluated the visual realism of Omni6D in comparison to other datasets through a survey involving 70 human subjects. We randomly selected 10 images from Omni6D, CAMERA~\cite{NOCS}, REAL~\cite{NOCS}, and Wild6D datasets~\cite{reponet}. To introduce noise, we blended in 2 images from COCO~\cite{coco}, which includes captured photos, and 3 images from SKETCH\footnote{https://sketchfab.com/}, which comprises rendered images. We randomly shuffled the order of the aforementioned 45 images and asked subjects to rate them anonymously, \ie, participants were unaware of the dataset to which each image belonged.
Subjects were asked to rate the realism of sampled images on a scale from 1 (least realistic) to 5 (most realistic). 
Here is the specific instruction for this survey: \textit{In this subsection, participants are required to rate the fidelity of the images, i.e., how closely they resemble images seen by the human eye. Ratings range from 1 to 5, with 1 representing a complete absence of fidelity and 5 denoting full congruence with perceptual images.}

\subsection{Questionnaire results}
\begin{figure}[t]
  \centering
   \includegraphics[width=\linewidth]{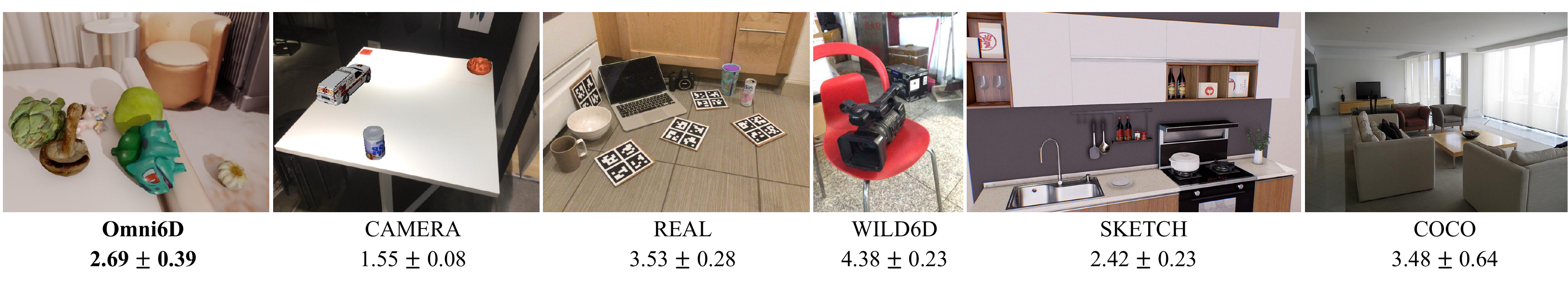}
   \caption{\textbf{Comparison of Visual Realism.} Complete results, including ratings for all datasets in the survey.}
   \label{fig:Qstar}
\end{figure}

\begin{figure}[t]
  \centering  
   \includegraphics[width=0.7\linewidth]{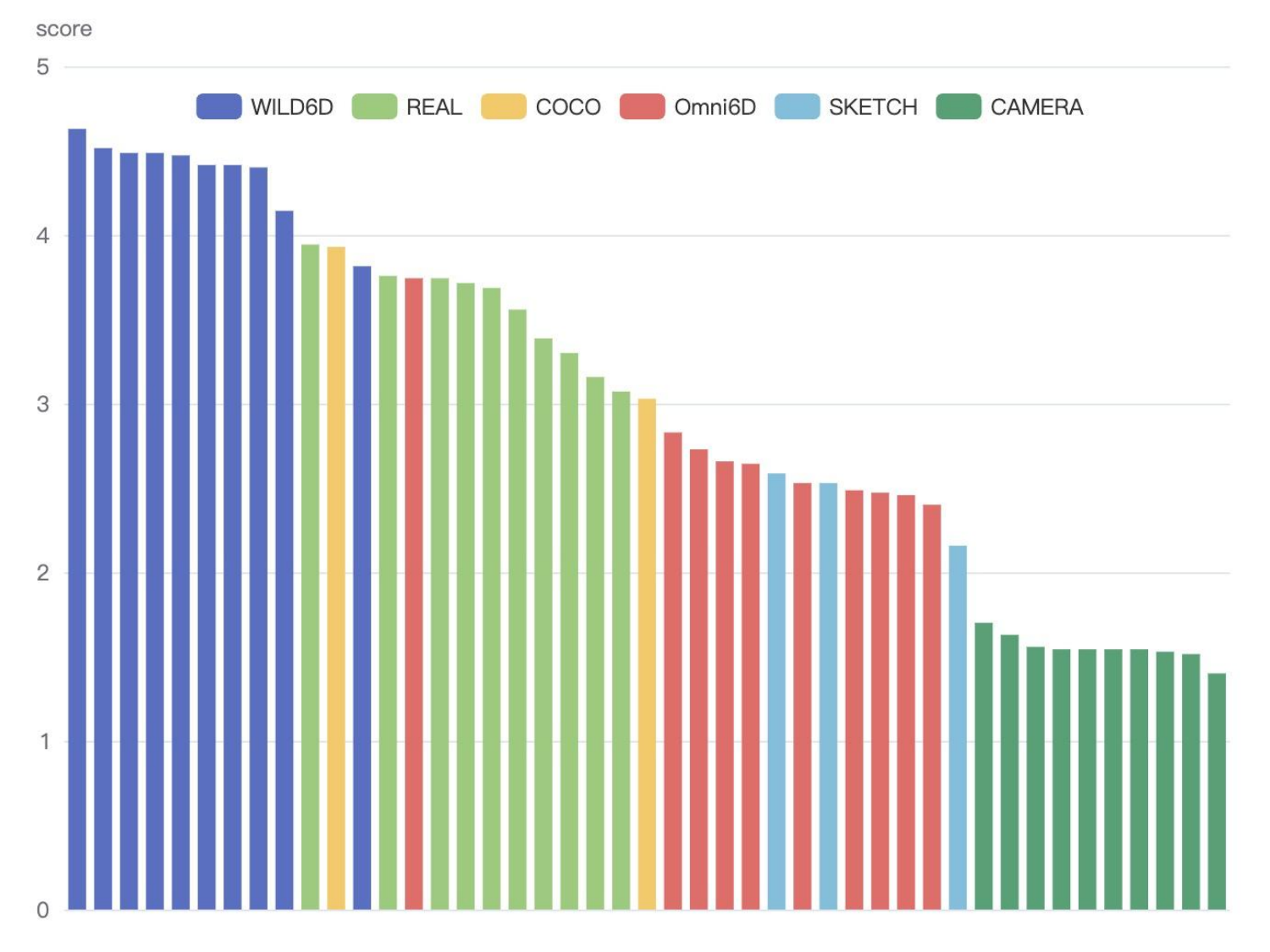}
   \caption{\textbf{Fidelity ratings for each image.} It displays the average ratings of all images in the questionnaire across 70 surveys, while the bar chart shows a gradual decrease in ratings from left to right, with each color representing a different dataset.}   
   \label{fig:score}
\end{figure}
We reported the average ratings and standard deviations for all datasets in \cref{fig:Qstar}, along with a sampled image from the questionnaire. \cref{fig:score} illustrates the average rating for each image. 
It can be observed that despite Omni6D having lower fidelity compared to captured photos, its ratings are significantly higher than those of CAMERA, which are also synthetic images. Furthermore, there is a noticeable gap between the ratings of Omni6D and CAMERA, with some images from Omni6D closely resembling captured photos.
\begin{figure}[t]
  \centering
  \begin{subfigure}{0.491\linewidth}
    \centering
    \includegraphics[width=\linewidth]{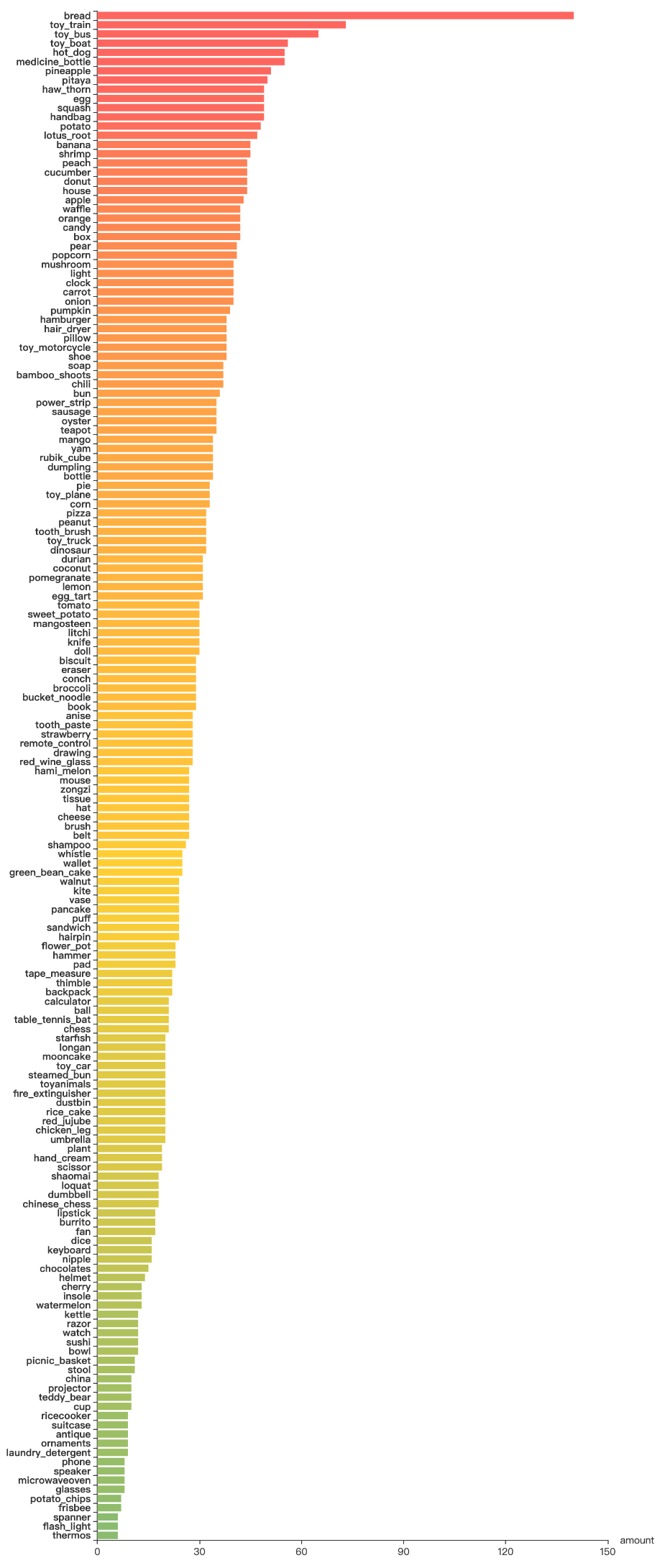}
    \caption{Instance count of category}
    \label{fig:cate_166}
  \end{subfigure}
  \begin{subfigure}{0.50\linewidth}
    \centering
    \includegraphics[width=\linewidth]{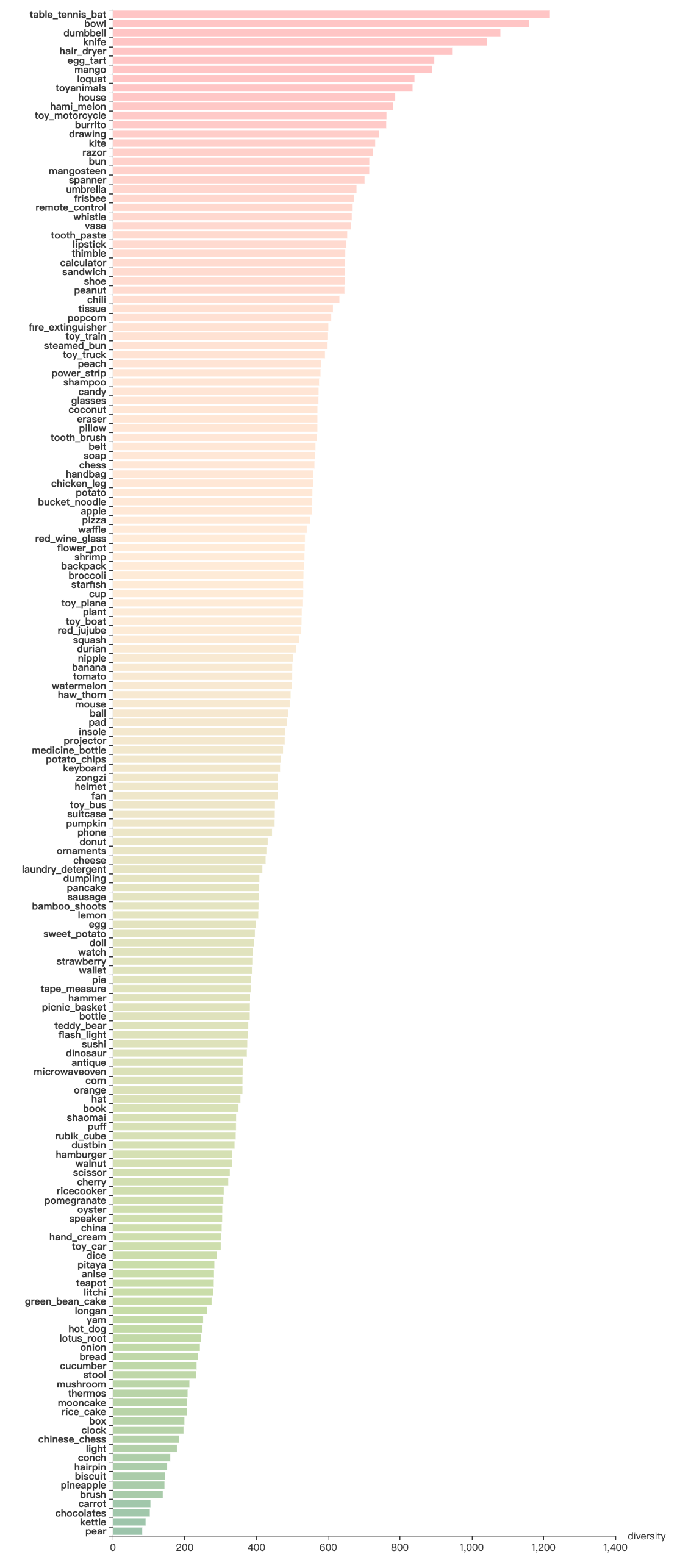}
    \caption{Intra-class diversity of category}
    \label{fig:diversity}
  \end{subfigure}
  \caption{\textbf{Omni6D Statistics.} \textbf{(a)} Category inventory and instance counts within Omni6D. Bars are sorted in descending order based on the instance counts of each category in the entire Omni6D dataset (train/val/test). \textbf{(b)} Intra-class diversity within categories in Omni6D. We measure the diversity of instances within a category using the mean Chamfer distance of all pairwise pairs within that category. Bars are sorted in descending order based on the intra-class diversity of each category in Omni6D$_{train}$.}
  \label{fig:all_cate}
\end{figure}

\begin{figure}[t]
  \centering
   \includegraphics[width=\linewidth]{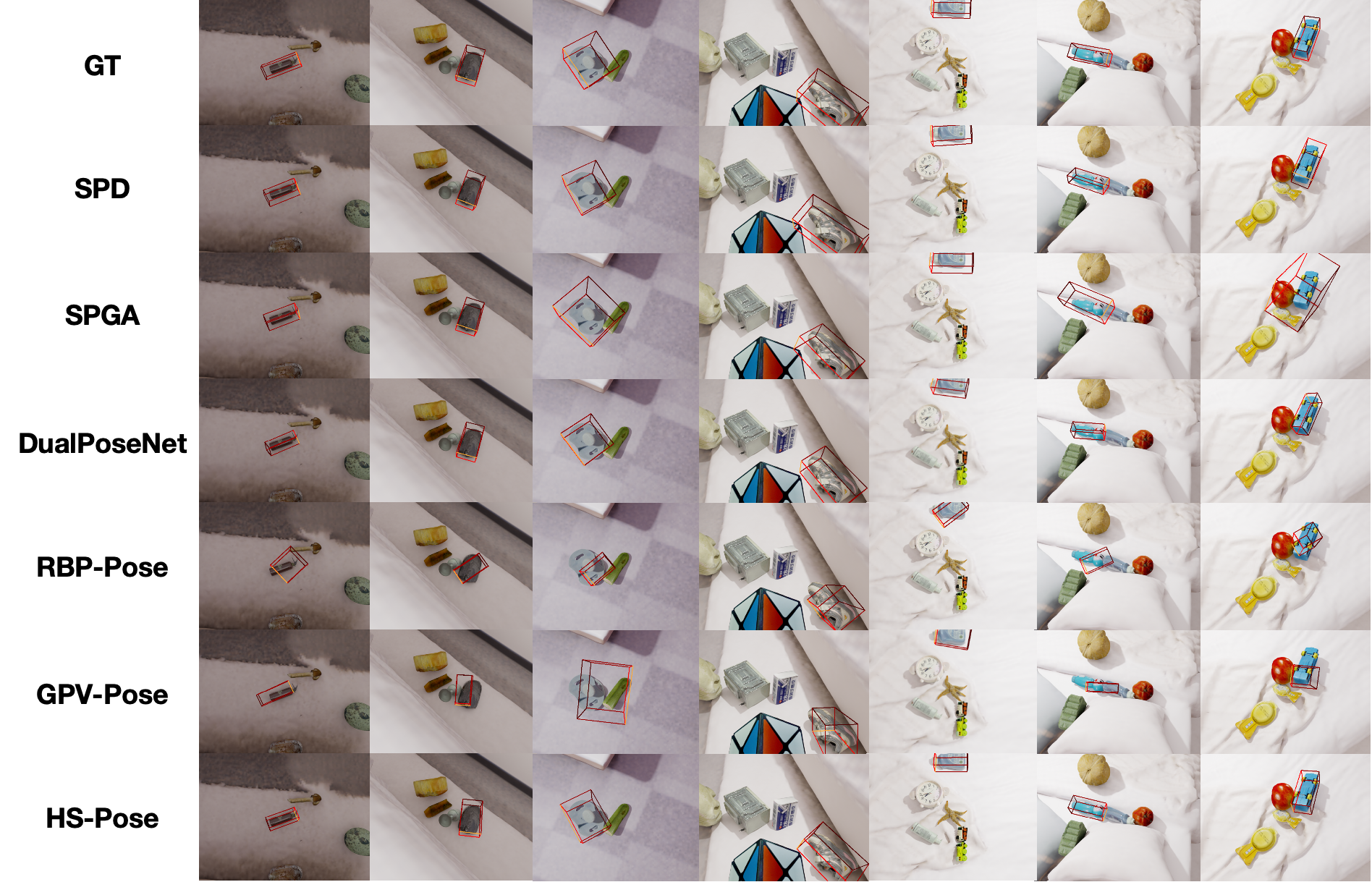}
   \caption{\textbf{Qualitative 6D pose and size estimation on Omni6D.} 
    From top to bottom, figures correspond to results of ground truth, SPD~\cite{SPD}, SGPA~\cite{SGPA}, DualPoseNet~\cite{DualPoseNet}, RBP-Pose~\cite{RBP}, GPV-Pose~\cite{GPV}, HS-Pose~\cite{HS} on Omni6D$_{test}$.}
   \label{fig:qualitative}
\end{figure}
\begin{figure}[t]
  \centering
   \includegraphics[width=\linewidth]{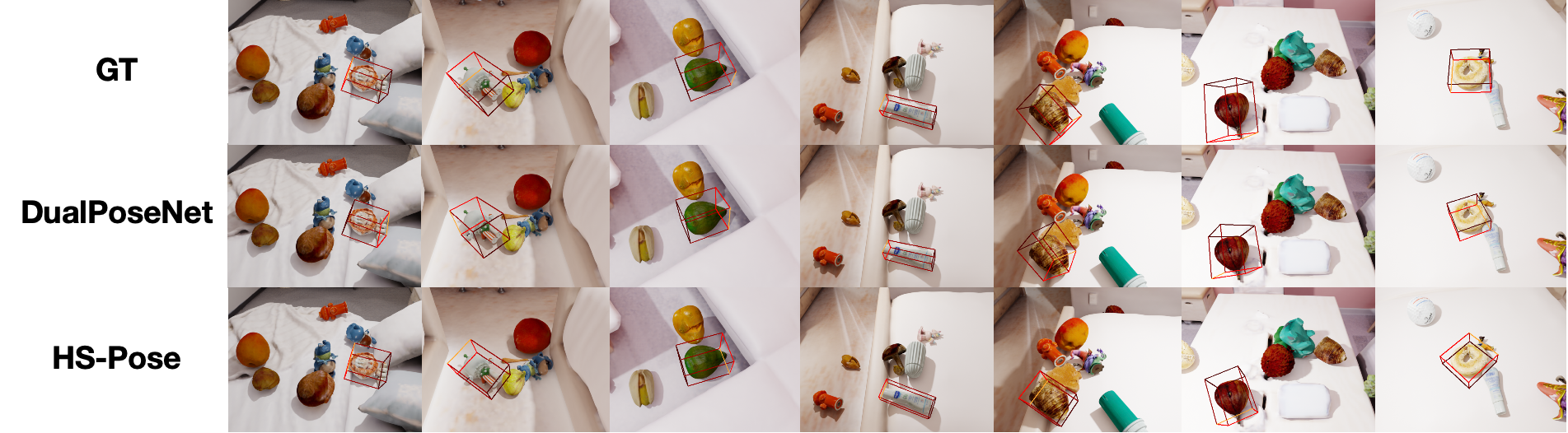}
   \caption{\textbf{Qualitative 6D pose and size estimation on unseen categories.}  From top to bottom, figures correspond to results of ground truth, DualPoseNet~\cite{DualPoseNet} and HS-Pose~\cite{HS} on Omni6D$_{out}$. We only showcase results from two models, DualPoseNet and HS-Pose, both of which exhibit inter-class generalization abilities.}
   \label{fig:qualitative_unseen}
\end{figure}


%
%


\section*{Acknowledgements}
\noindent This project is funded by ShanghaiAI Laboratory (P23KS00010.2022ZD0160201), the Centre for Perceptual and Interactive Intelligence (CPII) Ltd under the Innovation and Technology Commission (ITC)'s InnoHK, the Ministry of Education, Singapore, under its MOE AcRF Tier 2 (MOET2EP20221- 0012), NTU NAP, and under the RIE2020 Industry Alignment Fund – Industry Collaboration Projects (IAF-ICP) Funding Initiative.

\bibliographystyle{splncs04}
\bibliography{main}
\end{document}